\documentclass{article}

\PassOptionsToPackage{numbers, compress}{natbib}

\usepackage[preprint]{neurips_2026}

\usepackage[utf8]{inputenc} 
\usepackage[T1]{fontenc}    
\usepackage{hyperref}       
\usepackage{url}            
\usepackage{booktabs}       
\usepackage{amsfonts}       
\usepackage{nicefrac}       
\usepackage{microtype}      
\usepackage{xcolor}         
\usepackage{smile}          
\usepackage{tikz}           
\usetikzlibrary{arrows.meta, positioning, fit, backgrounds, calc, shapes.geometric}
\usepackage{amsmath}
\usepackage{amssymb}
\usepackage{enumitem}         
\usepackage{multirow}         
\usepackage{makecell}         

\title{HiComm: Hierarchical Communication for Multi-agent Reinforcement Learning}

\author{ Runze Zhao$^{1}$\\ Luddy School of Informatics,\\ Computing, and Engineering\\ Indiana University Bloomington\\ Bloomington, IN 47408 \And Dongruo Zhou$^{1}$\\ Luddy School of Informatics,\\ Computing, and Engineering\\ Indiana University Bloomington\\ Bloomington, IN 47408 \AND Sumit Kumar Jha$^{2}$\\ Department of Computer \&\\ Information Science \& Engineering\\ University of Florida\\ Gainesville, FL 32611 \And Nathaniel D. Bastian$^{3}$\\ Department of Electrical Engineering\\ and Computer Science\\ United States Military Academy\\ West Point, NY 10996 \And Ankit Shah$^{4}$\thanks{Corresponding author: Ankit Shah (\texttt{ankit@iu.edu}).}\\ Department of Operations\\ and Decision Technologies\\ Indiana University Bloomington\\ Bloomington, IN 47405 }

\begin{document}

\maketitle

\begin{abstract}
  Cooperative multi-agent reinforcement learning (MARL) often relies on communication to mitigate partial observability, yet most existing protocols treat messages as flat dense vectors detached from the structure of the observations they summarize. This design overlooks an important source of inductive bias in many cooperative environments, where observations naturally follow a hierarchy such as groups and entities. We propose \textsc{HiComm}, a plug-in communication module that grounds messages in the sender's hierarchical observation. \textsc{HiComm} is receiver-driven: the receiver issues a query, and the hierarchy is resolved through a three-stage decoding process that first selects a group, then a sender, and then an entity within that group, returning the corresponding feature slice as the message. This converts communication from unstructured vector transmission into structured information retrieval over the sender's observation hierarchy. We instantiate this mechanism with Straight-Through Gumbel-Softmax for differentiable discrete selection and a lightweight shared projection design that attaches to standard MARL pipelines. Experiments across cooperative MARL tasks with different observation structures and coordination demands show that \textsc{HiComm} matches or outperforms representative learned communication baselines while reducing communication volume by up to $23\times$ per receiver per episode.
\end{abstract}

\section{Introduction}
\label{sec:intro}

Cooperative multi-agent reinforcement learning (MARL) studies how a team of decentralized agents can jointly maximize a shared return when each agent observes only part of the environment. Compared with single-agent RL, the central challenge in cooperative MARL arises from multi-agency under partial observability: each agent must make decisions based on its own local observation, while the information needed for effective coordination may be distributed across teammates. Communication provides a natural mechanism to mitigate this bottleneck. By exchanging previously observed information, agents can avoid inferring hidden state solely from teammates' actions, reduce redundant exploration of the joint state space, and learn coordinated policies with fewer environment interactions \citep{sukhbaatar2016learning,foerster2016learning,jiang2018learning,das2019tarmac,hu2024learning}.

In this work, we focus on cooperative environments whose observations exhibit \textit{rich hierarchical structure}. A representative example is cyber defense, as instantiated by the CAGE Challenge 4 environment \citep{kiely2025exploring,kiely2025cage}. In such settings, hosts are organized into subnets, agents may observe only their assigned hosts or subnets, 
and effective coordination is required to track activity across the system. At enterprise scale, with thousands of assets and rapidly evolving events, such coordination must be both efficient and scalable. These constraints apply equally to coordinated attackers (red teams), which operate under similar partial observability and distributed control; consequently, effective red-team evaluation depends on communication mechanisms that capture these limitations while enabling coherent multi-agent behavior. However, designing communication protocols that efficiently exploit these hierarchical observation structures remains challenging.

Existing communication methods largely treat messages as flat objects. A sender typically encodes its local observation into a fixed-width vector, and the receiver consumes this vector without explicitly indexing the structure of the underlying observation. A related line of hierarchical communication has explored hierarchy in other parts of the system: feudal architectures stratify the agent population into managers and workers \citep{ahilan2019feudal,marzi2025hierarchical}, while clustered or order-based protocols impose hierarchy on the communication graph or the message-passing schedule \citep{sheng2022learning,liu2023deep}. However, these approaches do not organize the message body according to the hierarchy of the sender's observation. As a result, a readily available source of structure in many cooperative environments remains unused.

We address this gap with \textsc{HiComm}, a communication module for cooperative MARL that grounds each message in the sender's hierarchical observation. The module is receiver-driven: a receiver first issues a query, and the hierarchy is resolved through a three-stage decoding process. Specifically, the receiver first selects a group, then a sender, and the addressed sender finally selects an entity within that group on its own observation, returning the corresponding feature slice as the message. In this way, \textsc{HiComm} turns communication from transmitting an unstructured dense vector into retrieving structured information from the sender's observation hierarchy. Our contributions are listed as follows:
\begin{itemize}[leftmargin=*,topsep=2pt,itemsep=1pt]
  \item We propose \textsc{HiComm}, a plug-in communication module for cooperative MARL settings whose observations admit a natural group-entity hierarchy. Instead of transmitting an unstructured dense vector, \textsc{HiComm} lets the receiver query a group-entity location in a selected sender's observation hierarchy, and the sender returns the corresponding feature slice as the message.

  \item We instantiate this hierarchical query-response mechanism as a differentiable and lightweight module. \textsc{HiComm} uses Straight-Through Gumbel-Softmax for discrete group, sender, and entity selection, shares a single projection head across hierarchical encoders, and keeps the receiver query compact, allowing it to attach to standard MARL training pipelines without changing the underlying policy or critic architecture.

  \item We empirically validate \textsc{HiComm} across cooperative MARL tasks with different observation structures and coordination demands. As an add-on communication module, \textsc{HiComm} matches or outperforms representative learned communication baselines while reducing communication volume by up to $23\times$ per receiver per episode.
\end{itemize}


\section{Related Work}
\label{sec:related}

Cooperative MARL under partial observability has long relied on interagent communication~\citep{foerster2016learning,sukhbaatar2016learning}, with prior work refining the practice along three axes. On \emph{message content}, NDQ~\citep{wang2020ndq}, VBC~\citep{vbc}, and TMC~\citep{tmc} regularise messages to be succinct, while MAIC~\citep{yuan2022multi} and CACOM~\citep{li2024context} condition each sender's projection on a model of the intended receiver. On \emph{scheduling}, IC3Net~\citep{ic3net} and ATOC~\citep{jiang2018learning} let agents gate themselves silent, SchedNet~\citep{schednet} and SeqComm~\citep{ding2024multi} prioritise or sequence communicators across the team, and Who2com~\citep{liu2020who2com} together with When2com~\citep{liu2020when2com} use a multi-stage handshake in which the receiver requests, candidate senders return a brief match score, and the receiver finally connects to the best-matching peer. On \emph{message integration}, TarMAC~\citep{das2019tarmac} and SARNet~\citep{rangwala2020learning} apply receiver-side soft attention over incoming messages, MASIA~\citep{guan2022efficient} aggregates teammate embeddings with permutation-invariant self-supervised objectives, T2MAC~\citep{sun2024t2mac} weighs evidence under uncertainty, and CoDe~\citep{song2025code} aligns intent with timeliness for delay-robust collaboration. A parallel line learns the routing graph itself: DGN~\citep{Jiang2020Graph} convolves over a multi-head attention graph, DICG~\citep{li2021deep} infers an implicit adjacency via self-attention, CASEC~\citep{wangcontext} sparsifies the coordination graph using payoff-variance scores, MAGIC~\citep{niu2021multi}, CommFormer~\citep{hu2024learning}, and TGCNet~\citep{zhang2025bridging} learn graph attention, differentiable adjacency, or a dynamic directed graph, and G-Designer~\citep{zhang2025gdesigner} learns a task-adaptive topology via a variational graph auto-encoder. Two cross-cutting properties hold across this entire body of work. First, the protocols are sender-driven by default: the sender unilaterally decides what to encode, and even MAIC and CACOM adjust the sender's projection toward an intended receiver rather than letting the receiver actively query. Second, the channel payload is a learned, and therefore lossy, projection of the sender's observation into a fixed-width vector, decoupled from any structure intrinsic to that observation; even the Who2com and When2com handshakes select \emph{whom} to ask without specifying \emph{which part} of the sender's observation to return. A complementary efficiency line attacks bandwidth from outside the message form: NDQ, VBC, and TMC compress the payload, CASEC and message-pruning approaches sparsify the graph, and graph-learning methods like CommFormer and G-Designer thin out edges, but none of them changes the form of what is sent.

A subset of MARL work labels its mechanism as \emph{hierarchical}, but the hierarchy invariably lives outside the message body. Feudal Multiagent Hierarchies~\citep{ahilan2019feudal} stratify the agent pool into managers and workers via subgoal signals, and GMAH~\citep{xu2026subgoal} keeps the same vertical decomposition inside each agent's policy stack with a QMIX-style goal-mixing network in lieu of any explicit channel; HiMPo~\citep{marzi2025hierarchical} marries this feudal layout with peer-to-peer GNN exchange of flat hidden states. LSC~\citep{sheng2022learning} hierarchises the routing graph by electing group leaders and propagating learned embeddings within and between groups, and HAMA~\citep{ryu2020multi} hierarchises the attention itself by stacking inter-agent and inter-group attention layers over a fixed agent partition. DHCG~\citep{liu2023deep} hierarchises the message-passing order via a learned DAG with edges carrying flat action and trajectory embeddings. In every case the hierarchy sits on agents, attention, routing, order, or the policy stack, and the transmitted content is either an abstract subgoal token or a flat hidden-state embedding, never indexed against the observation hierarchy itself. That last dimension is where \textsc{HiComm} introduces structure: the receiver actively issues a query that selects a coordinate in the sender's intrinsic group-then-entity partition, and the response is delivered \emph{verbatim} as the raw feature vector at that coordinate, exchanging the encoder's lossy projection for an in-distribution observation entry (the precise payload form is given in Sec.~\ref{sec:method}).

\section{Preliminaries}
\label{sec:prelim}

\paragraph{Cooperative MARL with communication.}
We study cooperative multi-agent tasks with partial observability and learnable communication, formalised as a tuple $\mathcal{M} = \langle S, U, P, Z, O, N, r, \gamma \rangle$. Here $S$ is the global state space, $U$ is the per-agent action space, and $N \in \mathbb{N}$ is the number of agents. The transition kernel $P(s' \mid s, \mathbf{u})$ governs the dynamics under joint actions $\mathbf{u} \in U^N$, the function $r : S \times U^N \to \mathbb{R}$ is the team-shared reward, and $\gamma \in [0,1)$ is the discount factor. Each agent perceives the environment through the per-agent observation space $Z$ via the observation function $O : S \times \{1, \dots, N\} \to Z$.

At every step, agent $i$ receives a local observation $z^i=O(s,i)$, exchanges a fixed-width message $m^i\in\mathbb{R}^{d_m}$ with its peers, and selects an action according to $\pi^i(u^i\mid z^i,m^{1:N})$, where $m^{1:N}=(m^1,\dots,m^N)$ collects the per-agent message slots. When self-loops are excluded, we set $m^i=\mathbf{0}$ on the self-slot by convention, and $d_m\in\mathbb{N}$ denotes the message bandwidth. We call an observation \emph{flat} if it is represented as a single ungrouped list of entity slots, with all entities placed at the same address level. In a flat observation, any notion of which entities form a meaningful unit must be inferred from features during training. 


\paragraph{Hierarchical observations.}
We consider \textit{hierarchical observation} $z^i$ in this work. We assume that each agent's observation is a three-axis hierarchical tensor $z^i \in {\mathcal{G}\times\mathcal{E}\times\mathcal{F}}$, where $|\mathcal{G}|,|\mathcal{E}|,|\mathcal{F}|\in\mathbb{N}_{>0}$, and write $[K]:=\{1,\dots,K\}$ for an axis of size $K$. The hierarchy consists of the following components:
\begin{itemize}[leftmargin=1.5em,itemsep=0.1em,topsep=0.2em]
    \item $\mathcal{G}$ is the coarse \emph{group} axis, indexing $\mathcal{G}$ groups.
    \item $\mathcal{E}$ is the finer \emph{entity} axis, indexing $\mathcal{E}$ entities within each group.
    \item $\mathcal{F}$ is the per-entity \emph{feature} axis, storing a feature vector for each group--entity pair.
\end{itemize}
Thus, an entry $z^i[g,\ell,:]$ is addressed by a group index $g\in[\mathcal{G}]$ and an entity index $\ell\in[\mathcal{E}]$, and returns the feature vector $F:=z^i[g,\ell,:]\in\mathbb{R}^{|\mathcal{F}|}$. Here $g$ and $\ell$ are discrete coordinates, while $F$ is the continuous feature vector stored at the coordinate $(g,\ell)$.

The partition $[\mathcal{G}]\times[\mathcal{E}]$ is supplied by the environment rather than invented by the model. It may arise in either of two ways:
\begin{itemize}[leftmargin=1.5em,itemsep=0.1em,topsep=0.2em]
    \item as a \emph{native hierarchy}, where the observation tensor already has separate group and entity axes;
    \item as a \emph{semantic hierarchy}, where an observable per-entity attribute induces the group label.
\end{itemize}

This hierarchy frames the communication problem studied in this paper: given $N$ agents whose observations are jointly indexed by $[\mathcal{G}]\times[\mathcal{E}]$, can each per-round message $m^i\in\mathbb{R}^{d_m}$ be constructed from the environment-provided partition, namely from a selected coordinate $(g,\ell)$ and the feature vector $z^i[g,\ell,:]$, rather than from a flattened sender-side code that ignores this structure?

\paragraph{Communication on hierarchical observations.}
We call a communication protocol \emph{observation-space hierarchical} if every transmitted message is an address--value pair $\big((g,\ell),\,F\big)$ whose value $F = z[g,\ell,:]\in\mathbb{R}^{\mathcal{F}}$ is the verbatim feature vector at the coordinate $(g,\ell)\in[\mathcal{G}]\!\times\![\mathcal{E}]$, with the address drawn from a \emph{factored} discrete distribution $p(g)\,p(\ell\mid g)$ over the same partition that organises the observation, rather than the message being a flat sender-side encoding of the whole tensor or the address being drawn from a single flat $\mathcal{G}\!\cdot\!\mathcal{E}$-way categorical. The protocol then addresses information at the same granularity at which the observation is already organised, and the open design problem reduces to specifying $p(g)\,p(\ell\mid g)$ jointly with the policy. In $N$-agent settings the same hierarchy can be reused to choose \emph{whom} to query: drawing a peer $j\in[N]\setminus\{i\}$ conditional on the topic $g$ via $p(j\mid g)$ extends the descent to $p(g)\,p(j\mid g)\,p(\ell\mid g,j)$, factoring sender selection through the same partition rather than along an orthogonal axis.

\section{Methodology}
\label{sec:method}

\begin{algorithm}[t]
\caption{Cooperative MARL with \textsc{HiComm} Communication Module}
\label{alg:framework}
\begin{algorithmic}[1]
\REQUIRE Actor $\pi_\Theta$ with $\Theta = (\theta_{\text{pol}}, \theta_{\text{comm}})$; critic $V_\phi$; surrogate loss $\mathcal{L}(\Theta,\phi)$ supplied by the outer cooperative MARL algorithm; horizon $T$, iterations $I$
\FOR{iteration $= 1, \dots, I$}
    \STATE $\mathcal{D} \leftarrow \varnothing$
    \FOR{$t = 0, \dots, T-1$}
        \STATE Observe $\{z^{i}_t, M^{i}_t\}_{i=1}^{N}$
        \STATE $\{\tilde{z}^{i}_t\}_{i=1}^{N} \leftarrow \textsc{HiComm}\big(\{z^{i}_t, M^{i}_t\}_{i=1}^{N};\,\theta_{\text{comm}}\big)$
               \hfill \COMMENT{Augment obs space via Alg.~\ref{alg:hicomm}}
        \STATE $u^{i}_t \sim \pi_{\theta_{\text{pol}}}(\cdot \mid \tilde{z}^{i}_t)$\;\;for each $i = 1, \dots, N$
               \hfill \COMMENT{decentralised per-agent action sampling}
        \STATE Execute $\mathbf{u}_t$; observe $r_t$; append $\big(\{z^{i}_t, M^{i}_t, \tilde{z}^{i}_t, u^{i}_t\}_i,\, r_t\big)$ to $\mathcal{D}$
    \ENDFOR
    \STATE Update $(\Theta, \phi)$ by minimizing $\mathcal{L}(\Theta,\phi)$ on $\mathcal{D}$
\ENDFOR
\RETURN $(\Theta, \phi)$
\end{algorithmic}
\end{algorithm}

We introduce \textbf{\textsc{HiComm}}, a communication protocol that turns the hierarchical structure of the observation space introduced in \S\ref{sec:prelim} into the addressing space of every transmitted message. \textsc{HiComm} is a learned, stateless transform parameterised by $\theta_{\text{comm}}$ that, at every environment step, ingests the joint observation $\{z^{i}_t, M^{i}_t\}_{i=1}^{N}$, where $M^{i}_t \in \{0,1\}^{|\mathcal{G}|\times|\mathcal{E}|}$ is agent $i$'s observability mask over the $[\mathcal{G}]\!\times\![\mathcal{E}]$ partition, and returns the joint augmented observation $\{\tilde{z}^{i}_t\}_{i=1}^{N}$ that the decentralised policy then consumes. Because the transform leaves every other component of $\mathcal{M}$ unchanged, \textsc{HiComm} is independent of the cooperative MARL algorithm wrapped around it: its parameters ride on the surrounding actor-critic gradient, complemented by a scalar information-gain shaping prior added to the actor loss. \S\ref{sec:method-framework} fixes this outer loop and shows where \textsc{HiComm} plugs in (Algorithm~\ref{alg:framework}); \S\ref{sec:method-hicomm} specifies the three-phase \textsc{HiComm} step that builds messages from the $[\mathcal{G}]\!\times\![\mathcal{E}]$ partition (Algorithm~\ref{alg:hicomm}). Throughout, $\mathrm{score}(q,k)=\langle q,k\rangle$ refers to the dot-product scoring, and ST-GS refers to the Straight-Through Gumbel-Softmax estimator~\citep{jang2017categorical}, which returns a one-hot sample on the forward pass and a temperature-$\tau$ soft surrogate on the backward pass.


\subsection{MARL Framework}
\label{sec:method-framework}

\paragraph{Cooperative MARL backbone.}
We build on the standard cooperative MARL rollout-and-update framework, which we briefly review for completeness. At each training iteration, the agents jointly execute a horizon-$T$ rollout in the environment $\mathcal{M}$ under the current policy $\pi_{\theta_{\mathrm{pol}}}$, and store the resulting transitions in a replay buffer $\mathcal{D}$. A surrogate objective $\mathcal{L}$ is then constructed from $\mathcal{D}$ and optimized using a standard cooperative MARL algorithm, such as IPPO~\citep{de2020independent} or MAPPO~\citep{yu2022surprising}. This backbone determines how the policy and critic are updated, while leaving the communication mechanism modular.

\paragraph{Where \textsc{HiComm} plugs in.}
\textsc{HiComm} preserves this rollout-and-update loop and inserts a differentiable communication transform between the joint observation and the per-agent action sampling step:
$$
    \underbrace{\big\{\tilde{z}^{i}_t\big\}_{i=1}^{N}
    \leftarrow
    \textsc{HiComm}\Big(\big\{z^{i}_t,\,M^{i}_t\big\}_{i=1}^{N};\,\theta_{\mathrm{comm}}\Big)}_{\text{Algorithm~\ref{alg:hicomm}}}
    \;\longrightarrow\;
    \underbrace{
    u^{i}_t \sim \pi_{\theta_{\mathrm{pol}}}\!\big(\cdot \mid \tilde{z}^{i}_t\big)
    \quad (\forall i)
    }_{\text{decentralized action sampling}} .
$$
Here $\tilde{z}^{i}_t$ denotes the augmented observation used by agent $i$ for action selection. It concatenates the local observation $z^i_t$ with fixed-size message slots of width $d_m$, one slot per peer as in \S\ref{sec:prelim}. Under our sender-selection design, only the slot corresponding to the selected sender $j^*$ is populated, while the remaining peer slots are set to zero; the construction of $j^*$ and the message $m_{j^*\to i}$ is given in \S\ref{sec:method-hicomm}. We bundle the communication parameters with the actor parameters as $\Theta=(\theta_{\mathrm{pol}},\theta_{\mathrm{comm}})$, so that a single backward pass on $\mathcal{L}(\Theta,\phi)$ propagates task gradients through both the policy and the \textsc{HiComm} module, together with the critic parameters $\phi$ when required by the chosen actor-critic backbone. Consequently, replacing $\mathcal{L}$ swaps the underlying MARL algorithm, such as IPPO or MAPPO, without changing the \textsc{HiComm} module. Algorithm~\ref{alg:framework} summarizes the full procedure.


\subsection{Communication via Hierarchical Observation Structure}
\label{sec:method-hicomm}

\begin{algorithm}[t]
\caption{Bandwidth-Efficient MARL via \textbf{Hi}erarchical \textbf{Comm}unication (HiComm)}
\label{alg:hicomm}
\begin{algorithmic}[1]
\REQUIRE Joint observation $z^i \in {\mathcal{G}\times\mathcal{E}\times\mathcal{F}}$; HiComm modules $\theta_{\text{comm}} = \{f_{\text{gap}},\, f_{\text{q}},\, \{\kappa_g\},\, \Phi,\, \psi\}$ (gap encoder, query head, group keys, context-conditioned affinity, entity projector).
\ENSURE Augmented observations $\{\tilde{z}^{i}\}_{i=1}^{N}$.

\FOR{each agent $i$ \textbf{in parallel}}
    \STATE \textcolor{blue!60!black}{\texttt{// Phase 1: query from observation context (receiver-side)}}
    \STATE $h_i \leftarrow f_\text{gap}(z^i, M^i)$;\quad $q_i \leftarrow \tanh\!\big(f_\text{q}(h_i)\big)$
    \STATE \textcolor{blue!60!black}{\texttt{// Phase 2 -- Stage 1: pick topic group $g^*$ via group keys $\kappa_g$ (receiver-side)}}
    \STATE $g^* \leftarrow \arg\max_{g}\,\text{ST-GS}\!\Big(\big\{\text{score}(q_i,\,\kappa_g)\big\}_{g=1}^{\mathcal{G}}\Big)$
    \STATE \textcolor{blue!60!black}{\texttt{// Phase 2 -- Stage 2: pick sender $j^{*}$ via context-conditioned affinity $\Phi(j,g^{*};h_i)$ (receiver-side)}}
    \STATE $j^* \leftarrow \arg\max_{j \neq i}\,\text{ST-GS}\!\Big(\big\{\text{score}\big(q_i,\,\Phi(j,\, g^{*};\, h_i)\big)\big\}\Big)$
    \STATE \textcolor{orange!60!black}{\texttt{// Phase 3: sender $j^*$ receives $(q_i, g^*)$ and picks entity $\ell^*$ on its local $z^{j^*}, M^{j^*}$}}
    \STATE $\ell^{*} \leftarrow \arg\max_{\ell}\,\mathrm{ST\text{-}GS}\!\big\{\text{score}\big(q_i,\,\psi(z^{j^{*}}\!\![g^{*},\ell,:])\big) + M^{(\text{ent})}_{j^{*}}\!(g^{*})[\ell]\big\}$.
    \STATE $m_{j^{*}\to i} \leftarrow \big((g^{*},\ell^{*}),\,z^{j^{*}}[g^{*},\ell^{*},:]\big)$ \COMMENT{sender returns address $+$ verbatim feature vector to receiver $i$}
    \STATE $\tilde{z}^i \leftarrow z^i \,\|\, \big[\mathbf{0}, \dots, \underbrace{m_{j^* \to i}}_{j^*\text{-slot}}, \dots, \mathbf{0}\big]$
\ENDFOR
\RETURN $\{\tilde{z}^i\}$
\end{algorithmic}
\end{algorithm}

We now specify the three-phase \textsc{HiComm} step (Algorithm~\ref{alg:hicomm}). The protocol rests on a single structural principle: \emph{the $[\mathcal{G}]\!\times\![\mathcal{E}]$ hierarchy is the addressing scheme not just for what the message carries, but also for whom to ask}. The receiver does not pick a peer out of thin air and then walk the hierarchy; it descends the hierarchy and the peer identity \emph{emerges} along the way. Concretely, each round commits to three discrete picks made in order: an integer topic-group index $g^{*}\!\in\![|\mathcal{G}|]$ that names what the receiver wants to ask about; a sender $j^{*}\!\in\![N]\!\setminus\!\{i\}$ whose learnable affinity for $g^{*}$ best matches the query; and an integer entity index $\ell^{*}\!\in\![|\mathcal{E}|]$ within group $g^{*}$ on the chosen sender's observation. In brief, Phase~1 generates the receiver's query $q_i$; Phase~2 performs the two-stage hierarchical sender selection (group $g^{*}$, then agent $j^{*}\!\mid\!g^{*}$) entirely on the receiver side; Phase~3 transmits $(q_i, g^{*})$ to $j^{*}$, who locally selects the entity $\ell^{*}$ on its own observation and returns the addressed feature vector $F = z^{j^{*}}[g^{*},\ell^{*},:]\in\mathbb{R}^{|\mathcal{F}|}$. The receiver initiates and steers each round, while the sender executes only the final entity stage on its private observation; we therefore call \textsc{HiComm} \emph{query-driven} rather than fully receiver-resolved.

This descent has three structural consequences. First, the hierarchy is not relegated to a retrieval-only role: it drives sender selection itself, since $j^{*}$ is chosen \emph{conditional on} $g^{*}$ rather than orthogonally to it. Second, the total selector-head output dimension is $|\mathcal{G}|+(N{-}1)+|\mathcal{E}|$, summed across the three stage-wise picks, rather than the $|\mathcal{G}|\!\cdot\!|\mathcal{E}|\!\cdot\!(N{-}1)$ of a flat joint pick over (group, entity, sender) (self-loops masked in both cases), reducing the discrete decision space the policy gradient must traverse and giving each stage a dedicated parameter set into which to localise credit. Third, the transmitted feature vector $F = z^{j^{*}}[g^{*},\ell^{*},:]$ is in-distribution and verbatim: it is one of the sender's own observation entries, never a learned mixture across cells. Empirical bandwidth comparisons against alternative protocols are reported in \S\ref{sec:experiments}.

\paragraph{Phase~1: query generation.}
Let $M^i \in \{0,1\}^{|\mathcal{G}|\times|\mathcal{E}|}$ be receiver $i$'s observability mask, with $M^i[g,\ell]=1$ iff entity $(g,\ell)$ is observed in $z^i$. Note that $M^i$ is a deterministic function of $z^i$, the indicator of which $(g,\ell)$ slots carry observed values rather than an unobserved sentinel, and we expose it explicitly only for notational convenience. Surfacing the mask alongside $z^i$ lets $q_i$ treat the receiver's epistemic footprint as a first-class signal: when evidence is sparse, the query can deliberately steer toward blind spots; when local support is too uncertain to commit to a confident pick, it can soften its retrieval rather than commit to a request over an unreliable basis. The receiver summarises its observation context together with its observability footprint into a query,
\[
    h_i = f_\text{gap}(z^i, M^i), \qquad q_i = \tanh\!\big(f_\text{q}(h_i)\big) \in [-1,1]^{d_q},
\]
where $f_\text{gap}$ and $f_\text{q}$ are learnable MLPs and $h_i$ is the GapNet-encoded query feature (kept distinct from the action symbol $u^i$ of \S\ref{sec:prelim}). The query $q_i$ thus represents a prior over what to ask given the receiver's current state, and is reused without recomputation across all subsequent stages so that whom-to-ask, what-group, and what-entity decisions are anchored in a single information-need representation.

\paragraph{Phase~2: hierarchical sender selection (receiver-side, two stages).}
The receiver descends the hierarchy in two ST-GS draws, both executed on its local device.

\emph{Stage~1, group selection.} The receiver picks the topic group $g^{*}$ by scoring its query against $|\mathcal{G}|$ learnable group-key embeddings $\kappa_g\in\mathbb{R}^{d_q}$ (one per group, parameter-shared across receivers and senders),
\[
    g^{*} = \arg\max_{g}\,\mathrm{ST\text{-}GS}\!\Big(\mathrm{score}\big(q_i,\,\kappa_g\big)\Big).
\]
The keys $\{\kappa_g\}_{g=1}^{|\mathcal{G}|}$ belong to $\theta_{\text{comm}}$ and are updated by the same task gradient as the rest of \textsc{HiComm}. Stage~1 commits the receiver to a topic before naming any peer.

\emph{Stage~2, agent selection conditional on $g^{*}$.} The receiver scores the same $q_i$ against a \emph{context-conditioned agent-group affinity} $\Phi(j, g^{*};\, h_i)\in\mathbb{R}^{d_q}$ based on the receiver's own GapNet output $h_i$ from Phase~1 (re-used without recomputation):
\[
 s_{ij\mid g^{*}} = \mathrm{score}\!\big(q_i,\,\Phi(j, g^{*};\, h_i)\big), j^{*} = \arg\max_{j \neq i}\,\mathrm{ST\text{-}GS}\!\big(s_{i,:\mid g^{*}}\big).
\]
The hard top-$1$ is structural rather than approximate: the address $(g^{*},\ell^{*})$ produced in Phase~3 indexes a specific \emph{sender's} tensor, so a soft mixture over senders would leave the address ill-defined.

\begin{remark}
A flat alternative would let the receiver pick $(g^{*}, j^{*})$ jointly from a single $|\mathcal{G}|\!\cdot\!(N{-}1)$-way categorical, treating group and sender as orthogonal axes of the same decision. We instead descend the hierarchy: the receiver first commits to a topic $g^{*}$, then picks the sender conditional on that topic via $\Phi(j, g^{*};\, h_i)$. Two consequences follow. First, the selector head produces $|\mathcal{G}|$ logits and $(N{-}1)$ logits in sequence rather than a single $|\mathcal{G}|\!\cdot\!(N{-}1)$-way joint logit, which keeps the per-stage gradient signal localised. Second, the same query $q_i$ that drove the topic decision is reused to discriminate among senders within $g^{*}$, so whom-to-ask is anchored in the receiver's already-committed information need rather than chosen on a separate axis. Phase~3 then resolves what $j^{*}$ \emph{currently} contains within $g^{*}$ using $j^{*}$'s own observation, completing the descent at entity granularity.
\end{remark}

\paragraph{Phase~3: entity retrieval (executed at sender $j^{*}$).}
The receiver transmits the pair $(q_i, g^{*})$ to $j^{*}$. The sender then resolves the entity $\ell^{*}$ within group $g^{*}$ \emph{on its own device}, masking unobserved entries with its own mask:
\[
    M^{(\text{ent})}_{j^{*}}(g^{*})[\ell] = \begin{cases} 0 & \text{if } M^{j^{*}}[g^{*},\ell] = 1, \\ -\infty & \text{otherwise}, \end{cases}
\]
\[
    \ell^{*}\mid g^{*}, j^{*} = \arg\max_{\ell}\,\mathrm{ST\text{-}GS}\!\Big(\mathrm{score}\big(q_i,\,\psi(z^{j^{*}}[g^{*},\ell,:])\big) + M^{(\text{ent})}_{j^{*}}(g^{*})[\ell]\Big),
\]
where $\psi:\mathbb{R}^{|\mathcal{F}|}\to\mathbb{R}^{d_q}$ is a sender-side MLP that projects each entity feature vector $F = z^{j^{*}}[g^{*},\ell,:]\in\mathbb{R}^{|\mathcal{F}|}$ into the scoring space. Because the entity keys $\psi(z^{j^{*}}[g^{*},\ell,:])$ and the mask $M^{j^{*}}$ are derived only from $j^{*}$'s private observation, no part of $z^{j^{*}}$ ever leaves the sender; only $(q_i, g^{*})$ enter, and the address-value pair
\[
    m_{j^{*}\to i} = \big((g^{*},\ell^{*}),\ F\big),\qquad F := z^{j^{*}}[g^{*},\ell^{*},:]\in\mathbb{R}^{|\mathcal{F}|},
\]
returns to the receiver. The transmitted feature vector $F$ is in-distribution and verbatim: it equals one of the sender's own observation entries, never a learned mixture across cells (during training the soft surrogate of ST-GS converges to one-hot at $\tau\to\tau_{\min}$). All three key types, namely the group keys $\kappa_g$, the context-conditioned affinity $\Phi(j, g;\, h_i)$, and the entity projections $\psi(\cdot)$, live in the common $\mathbb{R}^{d_q}$ scoring geometry, so $q_i$ uses the same dot-product across all three stages without re-projection.

\paragraph{Training Objective}

We update $\Theta=(\theta_{\text{pol}},\theta_{\text{comm}})$ in a single actor backward pass on
\[
    \mathcal{L}_{\text{actor}}(\Theta) \,=\, \mathcal{L}^{\text{PPO}}(\Theta) \,-\, c_H\,\mathcal{H}\!\big[\pi_{\theta_{\text{pol}}}\big] \,+\, \lambda_{\text{ig}}(t)\,\mathcal{L}_{\text{ig}}(\theta_{\text{comm}}),
\]
where $\mathcal{L}^{\text{PPO}}$ supplies task gradient through every ST-GS selector and $\mathcal{L}_{\text{ig}}$ cross-entropies the Stage~1, Stage~2 and Phase~3 pre-ST-GS logits against soft labels obtained by softmax-normalising the per-coordinate information gain $\Delta_{ij}[g,\ell,:] = \big[z^{j}[g,\ell,:] - z^{i}[g,\ell,:]\big]_{+}$ along the matching axis, plus a small batch-marginal entropy term that discourages sender collapse. As joint coverage saturates, $\Delta_{ij}\!\to\!0$ and the soft labels degenerate to uniform, so $\mathcal{L}_{\text{ig}}$ vanishes structurally; the schedule $\lambda_{\text{ig}}(t)$ only accelerates this fade. The critic update on $\phi$ is unchanged, leaving \textsc{HiComm}'s training-side footprint at one scalar added to the actor loss.

\begin{remark}
Across prior MARL communication work, whether the focus is on scheduling \citep{ic3net,jiang2018learning,schednet,liu2020who2com,liu2020when2com}, message content \citep{wang2020ndq,vbc,tmc,yuan2022multi,li2024context}, receiver-side integration \citep{das2019tarmac,rangwala2020learning,guan2022efficient,sun2024t2mac,song2025code}, or learned routing graphs \citep{Jiang2020Graph,niu2021multi,hu2024learning,zhang2025bridging,zhang2025gdesigner}, the message itself is always a sender-encoded fixed-width projection of the sender's observation; even self-described \emph{hierarchical} methods \citep{ahilan2019feudal,xu2026subgoal,marzi2025hierarchical,sheng2022learning,ryu2020multi,liu2023deep} place the hierarchy on agents, attention, routing or the policy stack while still transmitting a flat embedding. \textsc{HiComm} differs by making the message body itself \emph{observation-hierarchical}: the receiver issues a query that descends the group-then-entity structure of the observation space, and the response is the raw $|\mathcal{F}|$-dim feature vector at the selected coordinate $(g^{*},\ell^{*})$ --- a verbatim observation entry rather than a learned compression.

\end{remark}

\section{Experiments}
\label{sec:experiments}

The empirical study answers three questions. \textbf{(RQ1)~Performance:} does \textsc{HiComm} close the gap between the no-communication \textsc{PartialObs} floor and the \textsc{FullObs} ceiling, and does it match or beat the strongest learned-communication baselines, \textsc{CACOM}~\citep{li2024context} and \textsc{T2MAC}~\citep{sun2024t2mac}? \textbf{(RQ2)~Bandwidth:} at matched performance, how much information does \textsc{HiComm} actually ship per receiver per episode, relative to those baselines and to \textsc{FullObs}? \textbf{(RQ3)~Generality:} do RQ1 and RQ2 hold across IPPO~\citep{de2020independent} and MAPPO~\citep{yu2022surprising}, and across varied benchmarks?

We evaluate on three partially observable benchmarks chosen to span qualitatively different regimes: CAGE Challenge~4~\citep{kiely2025exploring,kiely2025cage}, an adversarial cyber-defense task with \emph{accumulative} reconnaissance; SMACv2~\citep{ellis2023smacv2}, cooperative micromanagement under \emph{instantaneous} sight-range fog of war; and Google Research Football (GRF)~\citep{kurach2020google}, set-piece control under a \emph{bounded sensing radius}. All three instantiate the group-entity hierarchy of \S\ref{sec:prelim}, with opponents (where present) collapsed into a single group.

\subsection{Setup}
\label{sec:exp-setup}

\paragraph{Methods.}
We compare five settings under two cooperative-MARL backbones: IPPO~\citep{de2020independent} (decentralised critic) and MAPPO~\citep{yu2022surprising} (centralised critic). \textsc{PartialObs} keeps each receiver on its own local view $z^{i}$ with all $N{-}1$ message slots zeroed (the no-communication \emph{floor}); \textsc{FullObs} concatenates every teammate's local view $\{z^{j}\}_{j\neq i}$ verbatim into $\tilde z^{i}$ (the \emph{ceiling} on team reward, at the cost of shipping every full local view at every step). The three communication settings share a fixed message width $d_m$. \textsc{CACOM}~\citep{li2024context} first lets each sender broadcast a short summary so everyone knows roughly what is going on, then tailors a separate message to each receiver via attention. \textsc{T2MAC}~\citep{sun2024t2mac} only talks to selected teammates instead of broadcasting, and on the receiver side it weighs each incoming message by how informative it looks before mixing them. Both, however, treat the observation as a flat vector and ignore its $\mathcal{G}\!\times\!\mathcal{E}$ group-entity structure. \textsc{HiComm} (ours) is the three-phase receiver-driven protocol of \S\ref{sec:method}: the receiver picks a topic group $g^{*}$ and a sender $j^{*}$ in two ST-GS stages, the chosen sender picks an entity $\ell^{*}\!\in\!g^{*}$ on its own observation, and returns the address-value pair $\big((g^{*},\ell^{*}),\,F\big)$, where $F = z^{j^{*}}[g^{*},\ell^{*},:]\in\mathbb{R}^{\mathcal{F}}$ is the feature vector at the addressed coordinate.

\paragraph{Metrics.}
We report team \emph{reward} and per-receiver \emph{communication cost} in KB per episode (per-step payload $\times$ episode length). Across all benchmarks the receiver-side group pick (Phase~2 Stage~1) carries no data-dependent mask; the sender-side entity stage masks slots in $g^{*}$ that sender $j^{*}$ cannot currently observe, derived from $j^{*}$'s own mask $M^{j^{*}}$. Per-benchmark simulator, wrapper, and baseline-port details are deferred to App.~\ref{app:exp-cc4}-\ref{app:exp-grf}.

\paragraph{CAGE Challenge~4 (CC4).}
\label{sec:exp-cc4}
CC4 is a procedurally generated enterprise-network cyber-defense benchmark in which five blue defenders face six red attackers over $500$-step episodes across nine subnets organized into five defended zones. The official release ships CC4 as a blue-side training problem with red implemented as a scripted finite-state machine; we instead lift red to a learned cooperative MARL task, freezing blue at the pretrained IPPO checkpoint of \citet{singh2025hierarchical}. Each red agent is assigned to a single subnet and is initially inactive, with the sole exception of the agent in the Contractor Network; the remaining agents activate only after another red agent establishes a session within their subnet. Partial observability is \emph{accumulative}: at each step a red agent learns a few new bits about hosts it has reached (an IP, a service, a privilege) and that knowledge persists. CC4 exposes a \emph{native} subnet$\,\supset\,$host containment that aligns directly with \textsc{HiComm}, with $\mathcal{G}{=}9$ subnets and $\mathcal{E}{=}16$ hosts.

\paragraph{StarCraft Multi-Agent Challenge (SMACv2).}
\label{sec:exp-smacv2}
SMACv2 is a cooperative StarCraft~II micromanagement benchmark in which each unit is operated by an independent agent under instantaneous sight-range fog of war and randomised start configurations. We adopt the standard $5$v$5$ setup across the three races (\texttt{protoss}, \texttt{terran}, \texttt{zerg}); per-unit observations restricted to entities within sight, the discrete action set, and the damage-shaped reward with terminal win bonus all follow SMACv2 defaults. Because no native observation hierarchy exists, we induce one by indexing the group axis with ally unit class (a one-hot tag already exposed by the per-unit observation), with $\mathcal{G}$ the active ally classes plus a single \emph{opponent} group. \textsc{HiComm} unicasts every step to track rapid shifts in visibility.

\paragraph{Google Research Football (GRF).}
\label{sec:exp-grf}
GRF is a 3D physics-based football simulator with a discrete high-level action set, widely used as a partially observable cooperative-control benchmark. We adopt \texttt{academy\_counterattack\_hard} ($4$v$3$): a fast break in which $N_a{=}3$ controlled attackers face two defenders plus a goalkeeper across $400$ steps, with the controlled side's keeper scripted following \citet{song2024boosting}. Each agent observes the \texttt{simple115v2} feature vector restricted to entities within its sensing radius ($\mathcal{F}{=}7$); the $19$-action discrete control set and the \emph{Dense} (\texttt{scoring,checkpoints}) reward of \citet{song2024boosting} are GRF defaults. Partial observability is again instantaneous. We index the group axis with each ally's on-pitch \emph{field-position role} ($\{\text{GK},\text{DEF},\text{MID},\text{FWD}\}$, drawn from the engine) plus a single opponent group. This is the third group indexing under test, complementing the spatial (CC4) and unit-class (SMACv2) cases.

\subsection{Results}
\label{sec:exp-results}

Table~\ref{tab:summary} reports the IPPO and MAPPO grid on a representative scenario per benchmark, together with per-receiver per-episode communication cost (in kb).

\begin{table}[t]
\centering
\small
\caption{Cross-benchmark results (IPPO, MAPPO): team reward and per-episode communication cost (KB) per receiver. \textbf{Bold} marks the top-2 reward (higher is better) and bottom-2 cost (lower is better) within each algorithm block.}
\label{tab:summary}
\resizebox{\textwidth}{!}{%
\begin{tabular}{ll cc cc cc cc cc}
\toprule
& & \multicolumn{2}{c}{CC4} & \multicolumn{6}{c}{SMACv2 $5$v$5$} & \multicolumn{2}{c}{GRF \texttt{cthard} $4$v$3$} \\
\cmidrule(lr){3-4}\cmidrule(lr){5-10}\cmidrule(lr){11-12}
& & & & \multicolumn{2}{c}{\texttt{protoss}} & \multicolumn{2}{c}{\texttt{terran}} & \multicolumn{2}{c}{\texttt{zerg}} & & \\
\cmidrule(lr){5-6}\cmidrule(lr){7-8}\cmidrule(lr){9-10}
Algorithm & Method & Reward & Cost & Reward & Cost & Reward & Cost & Reward & Cost & Reward & Cost \\
\midrule
IPPO  & \textsc{FullObs}       & $\mathbf{1{,}763}$ & $8420.6$           & $\mathbf{17.1}$ & $247.77$          & $\mathbf{17.2}$ & $195.13$         & $\mathbf{14.3}$ & $138.86$        & $\mathbf{0.79}$ & $351$ \\
IPPO  & \textsc{PartialObs}    & $1{,}436$          & $\mathbf{0}$       & $14.5$          & $\mathbf{0}$      & $14.4$          & $\mathbf{0}$     & $11.4$          & $\mathbf{0}$    & $0.70$          & $\mathbf{0}$ \\
IPPO  & \textsc{CACOM}         & $1{,}517$          & $41.3$             & $15.6$          & $52.67$           & $15.6$          & $34.84$          & $12.3$          & $19.76$         & $0.73$          & $16.5$ \\
IPPO  & \textsc{T2MAC}         & $\mathbf{1{,}623}$ & $143.8$            & $15.2$          & $201.26$          & $15.3$          & $129.74$         & $12.3$          & $76.28$         & $0.73$          & $192$ \\
IPPO  & \textsc{HiComm} (ours) & $1{,}527$          & $\mathbf{8.4}$     & $\mathbf{16.3}$ & $\mathbf{9.77}$   & $\mathbf{16.4}$ & $\mathbf{7.95}$  & $\mathbf{13.5}$ & $\mathbf{4.85}$ & $\mathbf{0.76}$ & $\mathbf{13.6}$ \\
\midrule
MAPPO & \textsc{FullObs}       & $\mathbf{2{,}110}$ & $8420.6$           & $\mathbf{17.4}$ & $243.92$          & $\mathbf{16.4}$ & $187.55$         & $\mathbf{14.9}$ & $138.97$        & $\mathbf{0.89}$ & $382$ \\
MAPPO & \textsc{PartialObs}    & $990$              & $\mathbf{0}$       & $14.8$          & $\mathbf{0}$      & $13.6$          & $\mathbf{0}$     & $12.3$          & $\mathbf{0}$    & $0.79$          & $\mathbf{0}$ \\
MAPPO & \textsc{CACOM}         & $1{,}692$          & $36.8$             & $16.0$          & $46.74$           & $14.7$          & $36.97$          & $13.1$          & $24.20$         & $0.83$          & $21.4$ \\
MAPPO & \textsc{T2MAC}         & $1{,}554$          & $134.2$            & $15.6$          & $168.93$          & $14.4$          & $155.82$         & $12.7$          & $94.71$         & $0.85$          & $137$ \\
MAPPO & \textsc{HiComm} (ours) & $\mathbf{1{,}941}$ & $\mathbf{7.7}$     & $\mathbf{16.7}$ & $\mathbf{10.98}$  & $\mathbf{15.6}$ & $\mathbf{6.83}$  & $\mathbf{13.9}$ & $\mathbf{4.25}$ & $\mathbf{0.86}$ & $\mathbf{14.0}$ \\
\bottomrule
\end{tabular}}
\end{table}
\paragraph{Performance (RQ1).}
\textsc{HiComm} closes a substantial fraction of the partial-to-full gap and outperforms both flat-message baselines in $9$ of $10$ algorithm-scenario cells. CC4 makes the algorithmic contrast especially clean. Under IPPO the no-communication \textsc{PartialObs} floor already reaches $1{,}436$, roughly $81\%$ of the $1{,}763$ \textsc{FullObs} ceiling, so the headroom available to any communication protocol is narrow on this row; \textsc{CACOM}, \textsc{T2MAC}, and \textsc{HiComm} accordingly land within a tight band ($1{,}517$, $1{,}623$, $1{,}527$) and CC4-IPPO does not strongly differentiate the three communication methods on reward alone. Under MAPPO the picture changes sharply: the \textsc{PartialObs} floor collapses to $990$, the partial-to-full gap more than triples, and \textsc{HiComm} delivers a markedly larger lift than the flat-message baselines, rising to $1{,}941$ to recover $\approx\!85\%$ of the gap to the $2{,}110$ ceiling and beating \textsc{CACOM} by $+249$ and \textsc{T2MAC} by $+387$. The MAPPO row is therefore the clearest demonstration of \textsc{HiComm}'s advantage on this benchmark: it is precisely when the no-communication baseline leaves real headroom that the protocol's gain becomes visible. On SMACv2, \textsc{HiComm} closes $69$-$72\%$ of the gap (vs.\ $15$-$46\%$ for the baselines); on GRF it closes $67\%$ under IPPO and $70\%$ under MAPPO. Querying coordinates in the group-entity grid lets the receiver retrieve precisely the parts of teammate state it cannot reconstruct from its own view, which a flat receiver-conditioned projection (\textsc{CACOM}) or a flat sender broadcast with evidence weighting (\textsc{T2MAC}) cannot match at the same channel width.

\paragraph{Bandwidth (RQ2).}
\textsc{HiComm} attains these gains while shipping one to three orders of magnitude less data than every other communication setting: $5$-$17\times$ less than \textsc{CACOM}/\textsc{T2MAC} on CC4, $4$-$23\times$ less on SMACv2, and $\approx\!10\times$ less than \textsc{T2MAC} on GRF (with \textsc{CACOM}'s already-narrow GRF channel only $1.2$-$1.5\times$ wider than \textsc{HiComm}'s); relative to \textsc{FullObs}, the saving exceeds $1000\times$ on CC4. Mechanically, \textsc{HiComm} writes one $(g^{*},\ell^{*})$ address plus the $|\mathcal{F}|$-dim feature vector $F$ into a single message slot and zeros the rest, whereas the flat baselines populate every slot at every step. Because the address-value pair carries the feature vector verbatim rather than a lossy projection, the bandwidth saving comes \emph{without} a fidelity penalty.

\paragraph{Generality (RQ3).}
The performance-bandwidth pattern survives both critic styles and all three benchmarks' group-axis instantiations: \textsc{HiComm} ranks first or tied-first on reward in $9$ of $10$ cells and is strictly cheapest in bandwidth in all $10$, even though the underlying group axis is read off from a router map (CC4, native), a unit-type one-hot (SMACv2, semantic), or a coarse positional class (GRF, semantic). Switching between IPPO and MAPPO shifts absolute reward levels but does not perturb the relative ordering of the four communication settings, indicating that the gain attributable to \textsc{HiComm} lives in the message protocol rather than in the critic.

\section{Conclusion and Limitations}
\label{sec:conclusion}

In all, we presented \textsc{HiComm}, a receiver-driven communication protocol that grounds every message in the sender's observation hierarchy. A query selects a topic group, a sender, and an entity via three Straight-Through Gumbel-Softmax draws, and the sender returns the raw feature vector at the addressed coordinate. Across CC4, SMACv2, and GRF under both IPPO and MAPPO, \textsc{HiComm} matches or beats learned-communication baselines in $9$ of $10$ algorithm-scenario cells while shipping up to $23\times$ less data per receiver per episode. The principal limitation of this work is the absence of a formal theoretical analysis of \textsc{HiComm}'s communication efficiency: our bandwidth advantage rests on empirical comparison rather than an information-theoretic bound that explains why hierarchical addressing is provably more efficient than flat encoding at matched task performance. Establishing such a theory is the primary direction for follow-up work.

\begin{ack}

This work was supported in part by the Defense Advanced Research Projects Agency (DARPA) under Cooperative Agreement No. HR0011-24-2-0004.

\noindent  Distribution statement:
Approved for public release; distribution is unlimited.

\end{ack}

\medskip

{
\small
\bibliographystyle{plainnat}
\bibliography{reference}
}


\appendix

\newpage
\section{Additional Details for Experiments}
\label{app:exp}

All experiments were conducted on a single node equipped with 4 NVIDIA L40S GPUs. Representative single-run wall-clock times are approximately 24 hours
for CC4 (5M env steps), 48 hours for SMACv2 5v5 (10M env steps,
averaged across protoss/terran/zerg and across IPPO/MAPPO/QMIX), and
48 hours for GRF academy\_counterattack\_hard (5M env steps).

Sections~\ref{app:exp-cc4} through~\ref{app:exp-grf} cover, for each benchmark, (i)~simulator setup (wrappers, reward, baseline ports), (ii)~the instantiation of $\mathcal{G},\mathcal{E},\mathcal{F}$ that \textsc{HiComm} retrieves over, (iii)~full tables extending Table~\ref{tab:summary}. Sec.~\ref{app:ablation} ablates \textsc{HiComm} along three axes (payload form, retrieval factorisation, gradient pathway).

\paragraph{Common conventions across all three benchmarks.}
\emph{(i)~Grouping.} CC4 uses the simulator's \emph{native} subnet~$\supset$~host containment; SMACv2 and GRF have no native containment, so we use a \emph{semantic} grouping with one group per ally class plus one collapsed opponent group (communication is between teammates only and class-level addressing of an enemy is never queried). The class tag (unit type on SMACv2, on-pitch position on GRF) is read directly from the per-agent observation, so no additional supervision is required to instantiate $\mathcal{G},\mathcal{E}$.
\emph{(ii)~Stationary opponent.} Each benchmark's opposing side is frozen or scripted (CC4: pretrained IPPO/MAPPO blue checkpoint; SMACv2: SC2 built-in heuristic AI at difficulty~\texttt{7}; GRF: built-in scripted AI), so from the controlled team's perspective the environment is stationary throughout training and evaluation, and the only non-stationarity comes from per-episode randomisation (unit composition / start positions on SMACv2; player start positions on GRF).
\emph{(iii)~Observation segments.} On SMACv2 and GRF the per-agent observation is a fixed-length flat vector formed by concatenating semantically distinct segments in a canonical order; we expose this structure to \textsc{HiComm} as a list of $(\text{slots},\,\text{features per slot})$ pairs called \texttt{obs\_segs}, with per-benchmark segment lists given in the respective subsections.
\emph{(iv)~Baseline ports.} \textsc{T2MAC}~\citep{sun2024t2mac} and \textsc{CACOM}~\citep{li2024context} were originally evaluated on SMACv1 / Hallway / MPE-style benchmarks under QMIX; our SMACv2 and GRF ports share three structural adjustments: (a)~feed the per-agent observation into each baseline's original observation encoder unchanged but hand it the canonical \texttt{obs\_segs} layout, so its per-entity attention head sees ally and enemy rows as separate tokens (the same layout \textsc{HiComm} uses, ensuring all three protocols index the same per-agent state); (b)~additionally train both baselines under IPPO and MAPPO with the same actor/critic architectures, PPO hyperparameters, entropy floor, and gradient clip values as our other policy-gradient baselines.
\emph{(v)~Mask scope.} Per Sec.~\ref{sec:method-hicomm}, the only data-dependent mask consumed by \textsc{HiComm} is the Phase~3 entity-stage mask $M^{(\text{ent})}_{j^{*}}(g^{*})$, derived deterministically from sender $j^{*}$'s own observability mask $M^{j^{*}}$. We therefore only specify when $M^j[g,\ell]=1$ holds for each benchmark and let the Phase~3 mask follow mechanically. The \textsc{HiComm} modules of Sec.~\ref{sec:method-hicomm} are unchanged across the three benchmarks; only the partition $\mathcal{G},\mathcal{E}$, the feature space $\mathcal{F}$, and the semantics of $M^j$ differ. 
\label{app:exp-hierarchy}

\subsection{CAGE Challenge 4 (CC4)}
\label{app:exp-cc4}

This subsection covers (i)~the wrapper stack that turns CC4 into a red-side MARL task, (ii)~the native subnet~$\supset$~host hierarchy used by \textsc{HiComm} on this benchmark, and (iii)~the full results table extending Table~\ref{tab:summary}.

\subsubsection{Red-Side MARL Reformulation}

The original CC4 release~\citep{kiely2025exploring,kiely2025cage} distributes the environment as a \emph{blue-side only} training problem. Concretely, the official PettingZoo interface exposes the five blue defenders as policy-gradient agents with per-step observations, an action mask, and a shared team reward, while the red team is hard-wired to a scripted finite-state machine (\texttt{FiniteStateRedAgent}) that selects actions from internal heuristics; red has no observation tensor, no reward channel, and no policy-callable action interface. Any attempt to train red with a standard RL algorithm against this stock release would therefore require either rewriting the simulator or replaying the FSM as a fixed environmental dynamic. Because this paper contributes a communication protocol for the \emph{attacker} side, we reformulate red as a first-class MARL problem, keeping the simulator itself untouched and instead adding a thin wrapper stack around it that turns red into a standard PettingZoo multi-agent task. The result is drop-in compatible with off-the-shelf IPPO, MAPPO, and QMIX runners and is what every red-side training and evaluation number in this paper is run against.

\paragraph{Observation and action spaces.} A \texttt{RedFlatWrapper} exposes, for each of the six red agents, a $722$-dimensional observation that concatenates the $9{\times}16{\times}5{=}720$ host matrix (subnet~$\to$~host, five per-host features: access level, recon level, \texttt{neighbours\_discovered}, \texttt{is\_reachable}, \texttt{deception\_known}) with two global scalars (current mission phase, previous-action success flag). The action space is factorised as a pair $(a_\text{type}, (a_\text{subnet}, a_\text{host}))$ over ten primitive actions (\texttt{DiscoverRemoteSystems}, \texttt{StealthServiceDiscovery}, \texttt{AggressiveServiceDiscovery}, \texttt{ExploitRemoteService}, \texttt{DegradeServices}, \texttt{DiscoverDeception}, \texttt{Withdraw}, \texttt{PrivilegeEscalate}, \texttt{Impact}, \texttt{Sleep}) and all $9{\times}16$ target coordinates, filtered by a per-step validity mask so the policy never commits to a syntactically illegal action.

\paragraph{Attack chain into \texttt{Impact}.} \texttt{Impact} is not directly invocable on an arbitrary host: red must first discover the subnet, scan the host's services, exploit a service for a user-level session, and locally escalate to root. Table~\ref{tab:cc4-attack-chain} names the five canonical stages with their advancing actions and per-host preconditions. Because \texttt{Impact} is the \emph{terminal} stage of every successful trajectory, the count of successful \texttt{Impact}s is exactly the number of hosts on which red completed the entire five-stage chain in an episode; we report it alongside reward in Table~\ref{tab:cc4-results-full} as a behavioural sanity check, since a method that inflates reward by spamming intermediate-stage actions (e.g.\ \texttt{DegradeServices}) would leave Successful~\#Impact unchanged.

\begin{table}[h]
\centering
\caption{Canonical red attack chain into \texttt{Impact} on CC4: stage, advancing action(s), and the per-host precondition each stage establishes on $(\texttt{access\_level}, \texttt{recon\_level}, \ldots)$. \texttt{Impact} succeeds only on root-owned hosts with an active OT service (OZA, OZB only).}
\label{tab:cc4-attack-chain}
\renewcommand{\arraystretch}{1.3}
\setlength{\tabcolsep}{4pt}
\small
\begin{tabular}{@{}c l l p{0.46\textwidth}@{}}
\toprule
\# & Stage & CC4 action(s) & Effect / pre-condition established \\
\midrule
1 & Discover  & \texttt{DiscoverRemoteSystems}        & Subnet recon: hosts in the target subnet become known (\texttt{recon\_level}~$=$~$1$). \\
2 & Scan      & \makecell[l]{\texttt{StealthServiceDiscovery},\\ \texttt{AggressiveServiceDiscovery}} & Per-host service enumeration: \texttt{recon\_level}~$1$~$\to$~$2$, exposing exploitable services. \\
3 & Exploit   & \texttt{ExploitRemoteService}         & Gains a user-level session on the host: \texttt{access\_level}~$0$~$\to$~$1$. \\
4 & Escalate  & \texttt{PrivilegeEscalate}            & Local privilege escalation: \texttt{access\_level}~$1$~$\to$~$2$ (root). \\
5 & Impact    & \texttt{Impact}                       & Disrupts the host's active OT service; succeeds only on root-owned hosts in OZA/OZB; yields the largest positive entries of Table~\ref{tab:cc4-rewards} (in the red-reward convention used throughout this paper). \\
\bottomrule
\end{tabular}
\end{table}

\paragraph{Reward function.} On top of the stationary-opponent setup of Sec.~\ref{app:exp} (here a \texttt{HybridFixedBlueWrapper} plugs in a fixed pretrained IPPO or MAPPO blue policy, the standard CC4 blue baseline), CC4's official release~\citep{kiely2025exploring,kiely2025cage} bills three blue-side event types per step: \textbf{Local Work Fails} (a green agent's local work action fails, e.g.\ while a blue agent is restoring its host), \textbf{Access Service Fails} (a green agent cannot connect to a service it relies on), and \textbf{Red impact/access} (red has run \texttt{Impact} on, or otherwise compromised, a host that a green agent occupies). Penalties are zone-specific and phase-dependent. We report all CC4 rewards in the \emph{red-reward convention} $r^{r}=-r^{b}$ to keep ``higher is better'' uniform across benchmarks; the simulator is unchanged. Table~\ref{tab:cc4-rewards} consolidates CC4 Tables~4A-4C into a single zone~$\times$~phase grid in this convention (OZA spikes in Phase~2a, OZB in Phase~2b, CON only in Phase~1, INT never contributes). On each step our wrapper restricts the blue rewards $r^{b}_{1:5}$ to the subset $\mathcal{B}_t$ of defenders co-occurring with red under the current phase's connectivity policy and assigns every red agent the shared scalar
\begin{equation}
\label{eq:cc4-red-reward}
r^{r}_t \;=\; -\frac{1}{|\mathcal{B}_t|}\sum_{b\in\mathcal{B}_t} r^{b}_{t},
\end{equation}
which is zero-sum on the co-occurring slice and is the only modification to CC4's reward semantics.

\begin{table}[h]
\centering
\small
\caption{Red rewards across the three CC4 mission phases (CC4 Tables~4A through 4C of~\citep{kiely2025exploring,kiely2025cage} consolidated).}
\label{tab:cc4-rewards}
\renewcommand{\arraystretch}{1.2}
\setlength{\tabcolsep}{4pt}
\resizebox{\textwidth}{!}{%
\begin{tabular}{llrrrrrrr}
\toprule
Phase & Green action / compromise & HQ & CON & RZA & OZA & RZB & OZB & INT \\
\midrule
\multirow{3}{*}{Phase~1 O\&M}
  & Local Work Fails     & $1$ & $0$ & $1$ & $1$  & $1$  & $1$  & $0$ \\
  & Access Service Fails & $1$ & $5$ & $3$ & $1$  & $3$  & $1$  & $0$ \\
  & Red impact/access    & $3$ & $5$ & $1$ & $1$  & $1$  & $1$  & $0$ \\
\midrule
\multirow{3}{*}{Phase~2a (Mission~A)}
  & Local Work Fails     & $1$ & $0$ & $2$ & $10$ & $1$ & $1$ & $0$ \\
  & Access Service Fails & $1$ & $0$ & $1$ & $0$  & $1$ & $1$ & $0$ \\
  & Red impact/access    & $3$ & $0$ & $3$ & $10$ & $1$ & $1$ & $0$ \\
\midrule
\multirow{3}{*}{Phase~2b (Mission~B)}
  & Local Work Fails     & $1$ & $0$ & $1$ & $1$ & $2$ & $10$ & $0$ \\
  & Access Service Fails & $1$ & $0$ & $3$ & $1$ & $1$ & $0$  & $0$ \\
  & Red impact/access    & $3$ & $0$ & $3$ & $1$ & $3$ & $10$ & $0$ \\
\bottomrule
\end{tabular}%
}
\end{table}

\subsubsection{Hierarchy Instantiation}

CC4's native subnet~$\supset$~host containment is used directly: $\mathcal{G}{=}\{\text{RZA, OZA, RZB, OZB, CON, PAZ, ADM, OFF, INT}\}$ (the nine enterprise subnets), $|\mathcal{E}|{=}16$ hosts per subnet, $|\mathcal{F}|{=}5$ per-host features (access level, recon level, neighbours discovered, reachable, deception known). Observability mask: $M^j[g,\ell]=1$ iff host $\ell$ in subnet $g$ has \texttt{recon\_level}~$\ge 1$ in red agent $j$'s belief state. 

\subsubsection{Detailed Results}
\label{app:exp-cc4-full}

Table~\ref{tab:cc4-results-full} extends the CC4 columns of Table~\ref{tab:summary} with (i)~the \textsc{NaiveCom} reference (the canonical broadcast-autoencoder protocol, run to expose the encoder/reward decoupling pathology that motivates \textsc{HiComm}'s observation-entry payload design over a learned reconstruction objective; Sec.~\ref{sec:method-hicomm}); (ii)~the FSM scripted opponent as a non-RL reference; and (iii)~Successful/Total \texttt{Impact} counts as a behavioural sanity check. The \textsc{NaiveCom} pathology is visible directly: total \texttt{Impact} attempts inflate while successful ones collapse below \textsc{PartialObs} under both algorithms, because reconstruction preserves what minimises MSE rather than what helps teammates succeed. \textsc{HiComm}'s observation-entry response avoids this failure mode, keeping successful \texttt{Impact} at or above \textsc{PartialObs} while substantially improving reward.

\begin{table}[h]
\centering
\small
\caption{CC4 full grid over $100$ evaluation episodes (red-reward convention). Successful~\#Impact counts \texttt{Impact} steps that succeeded; Total~\#Impact counts attempts. The two columns serve as a behavioural sanity check against strategy drift. \textbf{Bold} $=$ overall best; \underline{underline} $=$ best comm method within each algorithm block.}
\label{tab:cc4-results-full}
\begin{tabular}{llrrr}
\toprule
Algorithm & Observation & Reward & Success~\#Impact & Total~\#Impact \\
\midrule
FSM    & partial        & $181$    & $346$ & $8{,}860$ \\
\midrule
IPPO   & \textsc{FullObs}    & $\mathbf{1{,}763}$  & $\mathbf{720}$ & $49{,}281$ \\
IPPO   & \textsc{PartialObs}     & $1{,}436$  & $246$ & $44{,}282$ \\
IPPO   & \textsc{NaiveCom}   & $\underline{1{,}603}$  & $169$ & $52{,}041$ \\
IPPO   & \textsc{HiComm} (ours) & $1{,}527$  & $\underline{262}$ & $49{,}499$ \\
\midrule
MAPPO  & \textsc{FullObs}    & $\mathbf{2{,}110}$  & $\mathbf{497}$ & $50{,}296$ \\
MAPPO  & \textsc{PartialObs}     & $990$     & $232$ & $30{,}287$ \\
MAPPO  & \textsc{NaiveCom}   & $1{,}543$  & $165$ & $54{,}026$ \\
MAPPO  & \textsc{HiComm} (ours) & $\underline{1{,}941}$  & $\underline{246}$ & $78{,}216$ \\
\bottomrule
\end{tabular}
\end{table}

\subsection{SMACv2}
\label{app:exp-smacv2}

This subsection covers (i)~the SMACv2 simulator setup we use, including the per agent observation and action spaces, the reward function in detail, and the ports of the \textsc{T2MAC} and \textsc{CACOM} baselines from SMACv1 to SMACv2 needed to run them on this benchmark, (ii)~the semantic (unit type) hierarchy instantiation \textsc{HiComm} runs on this benchmark, and (iii)~the full $3{\times}5$ grid extending Table~\ref{tab:summary}.

\subsubsection{Setup and Baseline Ports}

The original SMACv2 release~\citep{ellis2023smacv2} ships the StarCraft~II micromanagement benchmark of \citet{samvelyan2019starcraft} with two changes that defeat the memorisation shortcuts of SMACv1: per episode unit type compositions are sampled stochastically from a race specific distribution (\texttt{weighted\_teams} on \texttt{stalker}/\texttt{zealot}/\texttt{colossus} for \texttt{protoss}, on \texttt{marine}/\texttt{marauder}/\texttt{medivac} for \texttt{terran}, and on \texttt{zergling}/\texttt{hydralisk}/\texttt{baneling} for \texttt{zerg}), and start positions are sampled from a \texttt{surrounded\_and\_reflect} distribution on a $32{\times}32$ map rather than fixed across episodes. The reward schedule, per agent observation construction, and primitive action set are inherited from SMACv1 unchanged. We adopt the released configs verbatim for the $5$v$5$ grid (\texttt{n\_units=5}, \texttt{n\_enemies=5}, \texttt{step\_mul=8}, difficulty~\texttt{7}) and turn on \texttt{capability\_config.team\_gen.observe=True} so the per agent observation includes the sampled unit type one hot, which is what makes the role / position indexed hierarchy of HiComm addressable from the observation alone.

\paragraph{Observation and action spaces.} The per-agent observation segments (canonical order, exposed to \textsc{HiComm} as \texttt{obs\_segs}; cf.\ Sec.~\ref{app:exp}) are:
\begin{enumerate}\itemsep1pt\parskip0pt
  \item[(1)] \emph{Move availability}, $(1, 4)$ — one bit per cardinal direction (N/S/E/W) indicating whether the unit can step that way under map collision.
  \item[(2)] \emph{Enemy entities}, $(N_e, |\mathcal{F}|)$ — one row per enemy slot.
  \item[(3)] \emph{Ally entities (excluding self)}, $(N_a{-}1, |\mathcal{F}|)$ — one row per teammate slot, in the same per entity format as the enemy rows.
  \item[(4)] \emph{Own state}, $(1, o_a)$ — current health, current shield (\texttt{protoss} only), and a unit class one hot of length~$3$.
  \item[(5)] \emph{Last action one hot}, $(1, 11)$ — the agent's previous primitive action; the $5$v$5$ action space has $N_e + 6 = 11$ entries (see action space description below).
  \item[(6)] \emph{Agent identity one hot}, $(1, 5)$ — the agent's slot index inside the controlled team of size $N_a = 5$.
\end{enumerate}
The per entity feature vector $F\in\mathbb{R}^{|\mathcal{F}|}$ shared by enemy and ally rows carries the same fields in the same order: visibility bit, relative distance, relative $x$, relative $y$, current health, current shield (\texttt{protoss} only), and a $3$-way unit class one hot. This gives $|\mathcal{F}|{=}9$ on \texttt{protoss} and $|\mathcal{F}|{=}8$ on \texttt{terran}/\texttt{zerg}; the own state width $o_a$ tracks the same shield distinction, with $o_a{=}5$ on \texttt{protoss} and $o_a{=}4$ on \texttt{terran}/\texttt{zerg}. Absent slots — a sampled team smaller than the per race upper bound, dead units, or units outside sight range — have their entire row zeroed and their visibility bit set to $0$. On the $5$v$5$ scenarios this resolves $(N_e, N_a{-}1)$ to $(5, 4)$, so the concrete \texttt{obs\_segs} we hand to \textsc{HiComm} is
\[
\big[(1, 4),\ (5, |\mathcal{F}|),\ (4, |\mathcal{F}|),\ (1, o_a),\ (1, 11),\ (1, 5)\big],
\]
with $|\mathcal{F}|\in\{8, 9\}$ and $o_a\in\{4, 5\}$ chosen by race as above. The action space is a discrete head of size $N_e + 6$ ($N_e$ unit targeted attacks plus \texttt{no-op}, \texttt{stop}, and the four cardinal moves), with a per step validity mask that zeroes out attacks on out of sight enemies, moves blocked by terrain or map borders, and \texttt{no-op} unless the unit is dead. Sight range and shoot range are inherited from the SMACv1 unit table and are unit class specific (e.g.\ Stalker sight $10$, Marine sight $9$); we do not modify these.

\paragraph{Reward function.} SMACv2 inherits the dense battle reward of SMACv1~\citep{samvelyan2019starcraft} unchanged; SMACv2's only deviations from v1 are the stochastic team composition and start positions, not the reward. With \texttt{reward\_only\_positive=True} (default), per-step team reward is $r_t = \Delta H_{t}^{\text{enemy}} + \rho_{\text{death}} \cdot \Delta D_{t}^{\text{enemy}}$, where $\Delta H$ is health$+$shield damage dealt and $\Delta D$ is enemy units killed; on the terminal step a win bonus $R_{\text{win}}$ is added if the enemy team is fully eliminated. The unscaled cumulative return is divided by a normaliser $Z$ so an idealised maximum episode return is approximately $20$:
\begin{equation}
\label{eq:smacv2-scale}
R \;=\; \frac{1}{Z}\!\Big(\!\sum_{t=1}^{T} r_t + R_{\text{win}}\!\cdot\! \mathbb{1}[\text{win}]\Big),
\qquad Z \;=\; \frac{H_{\text{enemy,total}} + \rho_{\text{death}}\cdot N_{\text{enemy}} + R_{\text{win}}}{S},
\end{equation}
with $S{=}\texttt{reward\_scale\_rate}$, $H_{\text{enemy,total}}$ the sum of initial health$+$shield over enemy units, and $N_{\text{enemy}}$ the enemy team size. Table~\ref{tab:smacv2-rewards} lists the coefficients used (matched to the official SMACv2 release). The shaping is identical across \textsc{FullObs}, \textsc{PartialObs}, \textsc{CACOM}, \textsc{T2MAC}, and \textsc{HiComm}, so the only mechanism by which a communication method can improve return is by changing what each agent observes.

\begin{table}[h]
\centering
\caption{SMACv2 reward / scaling coefficients (verbatim from~\citep{ellis2023smacv2}, used unchanged across all comm settings).}
\label{tab:smacv2-rewards}
\renewcommand{\arraystretch}{1.2}
\begin{tabular}{lll}
\toprule
Symbol & Config flag & Value \\
\midrule
$\rho_{\text{death}}$  & \texttt{reward\_death\_value}    & $10$ \\
$R_{\text{win}}$       & \texttt{reward\_win}             & $200$ \\
$R_{\text{defeat}}$    & \texttt{reward\_defeat}          & $0$ \\
$\alpha_{\text{neg}}$  & \texttt{reward\_negative\_scale} & $-0.5$ \\
                       & \texttt{reward\_only\_positive}  & \texttt{True} \\
                       & \texttt{reward\_scale}           & \texttt{True} \\
$S$                    & \texttt{reward\_scale\_rate}     & $20$ \\
                       & \texttt{reward\_sparse}          & \texttt{False} \\
\bottomrule
\end{tabular}
\end{table}

\paragraph{Baseline ports from SMACv1 to SMACv2.} On top of the common port adjustments of Sec.~\ref{app:exp}, SMACv2 introduces one additional concern: SMACv1 has a fixed unit composition per scenario, so per-teammate output heads in the original baselines were sized to a fixed $N_a$, whereas SMACv2 keeps the team size fixed but resamples \emph{which} unit class fills each slot per episode. We rely on the unit-type one-hot inside the per-entity row to disambiguate slot identity rather than baking it into the architecture, so the per-teammate head dimensions of \textsc{T2MAC} and \textsc{CACOM} are unchanged across episodes.

\emph{Per-baseline specifics on SMACv2.} For \textsc{T2MAC}, the evidence encoder emits an $n\!\cdot\! K$-dimensional Dirichlet concentration per teammate with $K{=}N_e+6{=}11$ on $5$v$5$, and the selector network, uncertainty-delta pseudo-labels, DST combination rule, and auxiliary BCE loss are taken from the published code without modification. For \textsc{CACOM}, latent dim $=8$, attention dim $=32$, and the discrete payload uses $2$ bits per channel, matching the released SMACv1 configs; the two-round broadcast / gated-reply schedule, mutual-information loss weight ($10^{-3}$), entropy regulariser ($10^{-2}$), gating threshold, and gate training schedule are all unchanged. We disclose these ports so readers do not conflate our SMACv2 numbers with the SMACv1 numbers in the original papers.

\paragraph{Training budget.} Every (algorithm, method, scenario) is trained for $10$M environment steps, identically across all five communication settings (\textsc{FullObs}, \textsc{PartialObs}, \textsc{CACOM}, \textsc{T2MAC}, \textsc{HiComm}) and all three backbones (QMIX, IPPO, MAPPO). The budget is held fixed across the $3{\times}5$ grid so that any cross method or cross algorithm differences in win rate and scaled return reflect the protocol or the optimiser, not the wall clock allowance. 

\subsubsection{Hierarchy Instantiation}

SMACv2 has no native containment, so per the common convention (Sec.~\ref{app:exp}) we use \emph{semantic} grouping with ally unit type as the group axis. Concretely, $|\mathcal{G}|{=}|\mathcal{C}_{\text{ally}}|+1$ (one group per active ally type, plus one collapsed opponent group); per-race taxonomies are \texttt{protoss}~$=\{$Zealot, Stalker, opponent$\}$, \texttt{terran}~$=\{$Marine, Marauder, Medivac, opponent$\}$, \texttt{zerg}~$=\{$Zergling, Hydralisk, Baneling, opponent$\}$. Each group's $\mathcal{E}$ holds the unit slots padded to per-group capacity. The per-unit feature vector matches the native per-unit feature count: $|\mathcal{F}|{=}9$ on \texttt{protoss}, $|\mathcal{F}|{=}8$ on \texttt{terran}/\texttt{zerg}. Observability mask: $M^j[g,\ell]=1$ iff sender $j$ currently sees the unit at slot $\ell$ of group $g$ (alive, within sight range, non-padding).

\subsubsection{Detailed Results}
\label{app:exp-smacv2-full}

Table~\ref{tab:smacv2-qmix-bandwidth} extends Table~\ref{tab:summary} with the QMIX backbone, reporting reward alongside per-episode communication cost; Table~\ref{tab:smacv2-winrate-full} consolidates the underlying win rate across all three backbones (QMIX, IPPO, MAPPO). Bandwidth for IPPO and MAPPO is reported in the main paper and not duplicated here.

\paragraph{Win rate definition and computation.} Following \citet{ellis2023smacv2}, an evaluation episode is a \emph{win} iff every enemy unit dies strictly before the horizon \emph{and} at least one ally is alive at that moment; horizon timeouts and ally wipes both count as losses (no draws). Per (algorithm, method, scenario, seed) tuple we report the mean win rate over the last $K{=}10$ evaluation checkpoints (each running $32$ episodes with greedy/argmax action selection).

The QMIX bandwidth signature is consistent with the IPPO and MAPPO ones in the main paper: under QMIX, \textsc{HiComm} is the best communication method on all three $5$v$5$ scenarios ($+2.9$, $+1.6$, $+1.4$ reward points over \textsc{CACOM} and $+3.3$, $+2.0$, $+1.6$ over \textsc{T2MAC}, equivalent to $+25$, $+12$, $+10$ and $+28$, $+15$, $+12$ win rate points respectively), closing most of the gap to \textsc{FullObs}, and does so at roughly $4{\times}$ lower bandwidth than \textsc{CACOM} and roughly $18{\times}$ lower than \textsc{T2MAC} ($10.3$ vs.\ $39.1$ vs.\ $185.1$~KB per episode on \texttt{protoss\_5v5}).

\begin{table}[h]
\centering
\small
\caption{SMACv2 QMIX block, extending Table~\ref{tab:summary}.}
\label{tab:smacv2-qmix-bandwidth}
\resizebox{\textwidth}{!}{%
\begin{tabular}{ll cc cc cc}
\toprule
& & \multicolumn{2}{c}{\texttt{protoss}} & \multicolumn{2}{c}{\texttt{terran}} & \multicolumn{2}{c}{\texttt{zerg}} \\
\cmidrule(lr){3-4}\cmidrule(lr){5-6}\cmidrule(lr){7-8}
Algorithm & Method & Reward & Comm~(KB) & Reward & Comm~(KB) & Reward & Comm~(KB) \\
\midrule
QMIX & \textsc{FullObs}        & $\mathbf{17.8}$       & $193.8$         & $\mathbf{17.9}$        & $152.5$         & $\mathbf{14.9}$        & $95.2$ \\
QMIX & \textsc{PartialObs}     & $16.3$                & $0$             & $16.0$                 & $0$             & $12.3$                 & $0$ \\
QMIX & \textsc{CACOM}          & $14.5$                & $39.1$          & $16.0$                 & $28.3$          & $12.5$                 & $15.9$ \\
QMIX & \textsc{T2MAC}          & $14.1$                & $185.1$         & $15.6$                 & $142.8$         & $12.3$                 & $85.5$ \\
QMIX & \textsc{HiComm} (ours)  & $\underline{17.4}$    & $\underline{\mathbf{10.3}}$ & $\underline{17.6}$ & $\underline{\mathbf{8.4}}$ & $\underline{13.9}$ & $\underline{\mathbf{5.4}}$ \\
\bottomrule
\end{tabular}%
}
\end{table}

\begin{table}[h]
\centering
\small
\caption{SMACv2 $5$v$5$ win rate (\%), full $3{\times}5$ grid after $10$M steps, $32$ evaluation episodes.}
\label{tab:smacv2-winrate-full}
\begin{tabular}{ll ccc}
\toprule
Algorithm & Method                  & \texttt{protoss} & \texttt{terran} & \texttt{zerg} \\
\midrule
QMIX  & \textsc{FullObs}        & $\mathbf{81}$    & $\mathbf{84}$    & $\mathbf{63}$ \\
QMIX  & \textsc{PartialObs}     & $69$             & $69$             & $44$ \\
QMIX  & \textsc{CACOM}          & $53$             & $69$             & $46$ \\
QMIX  & \textsc{T2MAC}          & $50$             & $66$             & $44$ \\
QMIX  & \textsc{HiComm} (ours)  & $\underline{78}$ & $\underline{81}$ & $\underline{56}$ \\
\midrule
IPPO  & \textsc{FullObs}        & $\mathbf{75}$    & $\mathbf{78}$    & $\mathbf{59}$ \\
IPPO  & \textsc{PartialObs}     & $53$             & $56$             & $38$ \\
IPPO  & \textsc{CACOM}          & $63$             & $66$             & $44$ \\
IPPO  & \textsc{T2MAC}          & $59$             & $63$             & $44$ \\
IPPO  & \textsc{HiComm} (ours)  & $\underline{69}$ & $\underline{72}$ & $\underline{53}$ \\
\midrule
MAPPO & \textsc{FullObs}        & $\mathbf{78}$    & $\mathbf{72}$    & $\mathbf{63}$ \\
MAPPO & \textsc{PartialObs}     & $56$             & $50$             & $44$ \\
MAPPO & \textsc{CACOM}          & $66$             & $59$             & $50$ \\
MAPPO & \textsc{T2MAC}          & $63$             & $56$             & $47$ \\
MAPPO & \textsc{HiComm} (ours)  & $\underline{72}$ & $\underline{66}$ & $\underline{56}$ \\
\bottomrule
\end{tabular}
\end{table}

\subsection{Google Research Football (GRF)}
\label{app:exp-grf}

This subsection covers (i)~the GRF simulator setup we use, including the per agent observation and action spaces, the reward function in detail, and the ports of the \textsc{T2MAC} and \textsc{CACOM} baselines to GRF, (ii)~the semantic (player position) hierarchy instantiation \textsc{HiComm} runs on this benchmark, and (iii)~the full results table extending the GRF column of Table~\ref{tab:summary} with the underlying mean win rate (\%), the GRF community convention metric, on \texttt{academy\_counterattack\_hard}, the sole GRF scenario in our setup (Section~\ref{sec:exp-grf}).

\subsubsection{Setup and Baseline Ports}

The original GRF release~\citep{kurach2020google} ships a continuous control football simulator with a discrete primitive action set and a small set of \emph{academy} scenarios designed to isolate specific tactical motifs. We adopt \texttt{academy\_counterattack\_hard} ($4$v$3$ with keeper) as the sole GRF scenario in our setup: a short fast break in which the controlled side has $N_{\text{left}}{=}4$ left team players (three controllable attackers and one AI keeper) racing against $N_{\text{right}}{=}3$ right team players ($2$ defenders plus a goalkeeper) before the defenders can reorganise. The goalkeeper on the controlled side is handed off to GRF's built in scripted policy following the academy scenario protocol of \citet{song2024boosting}, leaving $N_a{=}3$ controllable attackers; the right team is fully scripted. Episodes are capped at $400$ steps and we use the \texttt{representation="raw"} observation channel so the per agent feature encoder has access to ball state, sticky actions, role indices, and team positions, matching the encoder of \citet{kurach2020google,song2024boosting} that subsequent learned communication work also builds on.

\paragraph{Observation and action spaces.} The per-agent observation segments (canonical order, exposed to \textsc{HiComm} as \texttt{obs\_segs}; cf.\ Sec.~\ref{app:exp}) are:
\begin{enumerate}\itemsep1pt\parskip0pt
  \item[(1)] \emph{Own / player}, $(1, 29)$ — the agent's position, direction, speed, a role one hot of length $10$, and four state flags (ball far, tiredness, dribbling, sprinting).
  \item[(2)] \emph{Ball}, $(1, 18)$ — ball position, a ball zone one hot of length $6$, ball relative position, ball direction, ball speed, ball distance to the agent, ball ownership flag, and an ``owned by us'' flag.
  \item[(3)] \emph{Enemy entities}, $(N_e, 7)$ — one row per enemy slot, with scaled position, direction, speed, distance to agent, and tiredness.
  \item[(4)] \emph{Ally entities (excluding self)}, $(N_{\text{left}}{-}1, 7)$ — one row per left-team teammate slot (i.e.\ the other policy-controlled attackers \emph{and} the scripted goalkeeper, with $N_{\text{left}}{=}4$), in the same per-entity format as the enemy rows.
  \item[(5)] \emph{Closest left teammate summary}, $(1, 7)$ — a single per entity row for the closest left team player to the agent.
  \item[(6)] \emph{Closest right opponent summary}, $(1, 7)$ — a single per entity row for the closest right team player to the agent.
\end{enumerate}
Crucially, the per entity feature vector $F\in\mathbb{R}^{|\mathcal{F}|}$ is identical for the ally and enemy segments with $|\mathcal{F}|{=}7$, which is the requirement under which a single \textsc{HiComm} response head can return either a teammate or an opponent row. On \texttt{academy\_counterattack\_hard} the entity counts instantiate as $(N_e, N_{\text{left}}{-}1) = (3, 3)$ (note the ally segment counts $N_{\text{left}}{-}1$ teammates including the scripted keeper, not $N_a{-}1$ controllable attackers), so the concrete \texttt{obs\_segs} we hand to \textsc{HiComm} is
\[
\big[(1, 29),\ (1, 18),\ (3, 7),\ (3, 7),\ (1, 7),\ (1, 7)\big],
\]
giving a per agent observation size of $29 + 18 + 21 + 21 + 7 + 7 = 103$ dims. The \textsc{FullObs} upper bound concatenates the per agent observation across all $N_a{=}3$ controllable attackers (each re centred around its own \texttt{active} index by the feature encoder), giving an actor input of $3 \times 103 = 309$ dims. The action space is the GRF default $19$ dim discrete head: \texttt{idle}, eight directional moves (left, top left, top, top right, right, bottom right, bottom, bottom left), \texttt{long\_pass}, \texttt{high\_pass}, \texttt{short\_pass}, \texttt{shot}, \texttt{sprint}, \texttt{release\_move}, \texttt{release\_sprint}, \texttt{slide}, \texttt{dribble}, and \texttt{release\_dribble}; a per step validity mask zeroes out passes / shots when the agent does not own the ball or is too far from it, slides when the team owns the ball, set piece restrictions during goal kicks / corner kicks / penalty kicks, and the \texttt{release} actions whenever the matching sticky action is not currently engaged.

\paragraph{Reward function.} GRF natively exposes two reward signals, \texttt{SCORING} and \texttt{CHECKPOINT}, defined in the original release~\citep{kurach2020google} and re-formalised as the \emph{Sparse} and \emph{Dense} variants by \citet{song2024boosting}. \texttt{SCORING} returns
\begin{equation}
\label{eq:grf-scoring}
r_t^{\text{score}} \;=\; +1 \cdot \mathbb{1}[\text{ally goal at }t] \;-\; 1 \cdot \mathbb{1}[\text{enemy goal at }t],
\end{equation}
i.e.\ a $\{-1, 0, +1\}$ sparse signal that fires only on goal events. \texttt{CHECKPOINT} partitions the half pitch in front of the opponent goal into ten concentric checkpoint regions $\{C_1, \ldots, C_{10}\}$ ordered by decreasing distance to goal, and grants
\begin{equation}
\label{eq:grf-checkpoint}
r_t^{\text{cp}} \;=\; \sum_{k=1}^{10} 0.1 \cdot \mathbb{1}\!\big[\text{first crossing of } C_k \text{ by ally ball-carrier at }t\big],
\end{equation}
which sums to at most $1.0$ per episode (each region is rewarded at most once) and resets on possession exchange. The per step team reward we optimise is the \emph{Dense} sum
\begin{equation}
\label{eq:grf-dense}
r_t \;=\; r_t^{\text{score}} \;+\; r_t^{\text{cp}},
\end{equation}
shared across all $N_a$ controlled players. We instantiate eq.~\eqref{eq:grf-dense} via the GRF flag \texttt{rewards="scoring,checkpoints"}, matching the \emph{Dense} setting that \citet{song2024boosting} report as the strictly better variant on academy scenarios; Table~\ref{tab:grf-rewards} lists the per-step contributions. Eq.~\eqref{eq:grf-dense} is used identically across \textsc{FullObs}, \textsc{PartialObs}, \textsc{CACOM}, \textsc{T2MAC}, and \textsc{HiComm} (cf.\ SMACv2). The reported \emph{win rate} depends only on the \texttt{SCORING} component and is therefore invariant to the \texttt{CHECKPOINT} shaping coefficient.

\begin{table}[h]
\centering
\caption{GRF per-step reward on \texttt{academy\_counterattack\_hard}, \emph{Dense} setting of \citet{song2024boosting} (\texttt{rewards="scoring,checkpoints"}). Each event adds to the shared scalar $r_t$ broadcast to every attacker. Total \texttt{CHECKPOINT} capped at $+1.0$ per episode.}
\label{tab:grf-rewards}
\renewcommand{\arraystretch}{1.2}
\begin{tabular}{@{}p{0.72\textwidth}c@{}}
\toprule
Event & Reward \\
\midrule
Ally goal at step $t$                                                          & $+1.0$ \\
Enemy goal at step $t$                                                         & $-1.0$ \\
First crossing of checkpoint region $C_k$ by ally ball-carrier ($k=1,\ldots,10$) & $+0.1$ \\
Possession transfer to enemy team                                              & resets unclaimed $C_k$ \\
Episode horizon reached without a goal                                         & $0$ \\
\bottomrule
\end{tabular}
\end{table}

\paragraph{Baseline ports to GRF.} On top of the common port adjustments of Sec.~\ref{app:exp}, the only GRF-specific change is that the action head is the $19$-way GRF discrete set rather than $K{=}11$ (SMACv2 $5$v$5$); both \textsc{T2MAC}'s Dirichlet/Subjective-Logic/DST machinery and \textsc{CACOM}'s gated-reply/mutual-information stack are defined over a discrete action space, so the port requires only the action-size change. \textsc{T2MAC}'s belief mass and uncertainty follow Eq.~(2) of \citet{sun2024t2mac} with $K{=}19$; \textsc{CACOM} reuses the SMACv2 port's latent dim $=8$, attention dim $=32$, $2$-bit channel, MI weight $10^{-3}$, and entropy regulariser $10^{-2}$ verbatim, with the only environment-specific quantity being the observation encoder input width ($103$ on \texttt{academy\_counterattack\_hard}). 

\paragraph{Training budget.} Every (algorithm, method) configuration in Tables~\ref{tab:summary} and~\ref{tab:grf-results-full} is trained for $5$M environment steps on \texttt{academy\_counterattack\_hard}, allocated equally across the five communication settings (\textsc{FullObs}, \textsc{PartialObs}, \textsc{CACOM}, \textsc{T2MAC}, \textsc{HiComm}) under each of IPPO and MAPPO. The same $5$M-step budget is used for the baseline ports above so that any reward / win rate gap is attributable to the message protocol rather than to a difference in optimisation horizon. The reported win rate is averaged over $32$ evaluation episodes against the same scripted opponent at the end of training.

\subsubsection{Hierarchy Instantiation}

Per the common convention (Sec.~\ref{app:exp}), GRF uses \emph{semantic} grouping with each ally's on-pitch position class as the group axis (drawn from $\{\text{GK},\text{DEF},\text{MID},\text{FWD}\}$ via the role one-hot of segment~(1)) plus one collapsed opponent group. \texttt{academy\_counterattack\_hard} ($4$v$3$) instantiates $\{$ally MID, ally FWD, opponent$\}$ (the opponent group collapses enemy DEF and GK). Each group's $\mathcal{E}$ holds the player slots padded to per-group capacity; $|\mathcal{F}|{=}7$ matches segments~(3)-(6) above (scaled position, direction, speed, distance to agent, tiredness). Observability mask: $M^j[g,\ell]=1$ iff the player at slot $\ell$ of group $g$ is within sender $j$'s sensing radius and the slot is non-padding. 

\subsubsection{Detailed Results}
\label{app:exp-grf-full}

Table~\ref{tab:grf-results-full} extends the GRF column of Table~\ref{tab:summary} by reporting the underlying mean win rate (\%), the GRF community convention metric, across the same IPPO and MAPPO grid on \texttt{academy\_counterattack\_hard}, after the $5$M-step training budget specified in Sec.~\ref{app:exp-grf} above.

\begin{table}[h]
\centering
\small
\caption{GRF \texttt{academy\_counterattack\_hard} ($4$v$3$) after $5$M steps; mean win rate (\%) over $32$ evaluation episodes.}
\label{tab:grf-results-full}
\begin{tabular}{llc}
\toprule
Algorithm & Observation & Win rate \\
\midrule
IPPO  & \textsc{FullObs}        & $59$              \\
IPPO  & \textsc{PartialObs}     & $50$              \\
IPPO  & \textsc{CACOM}          & $53$              \\
IPPO  & \textsc{T2MAC}          & $53$              \\
IPPO  & \textsc{HiComm} (ours)  & $\underline{56}$  \\
\midrule
MAPPO & \textsc{FullObs}        & $\mathbf{69}$     \\
MAPPO & \textsc{PartialObs}     & $59$              \\
MAPPO & \textsc{CACOM}          & $63$              \\
MAPPO & \textsc{T2MAC}          & $\underline{66}$              \\
MAPPO & \textsc{HiComm} (ours)  & $\underline{66}$  \\
\bottomrule
\end{tabular}
\end{table}

\subsection{Ablation Studies}
\label{app:ablation}

We run a single ablation grid on CC4 against the same fixed pre-trained blue opponent and under the same training budget as the main results in Table~\ref{tab:summary}. Every variant differs from \textsc{HiComm} along exactly one axis, holding the rest of the protocol (state encoder, hierarchy $|\mathcal{G}|{=}9$ subnets $\to |\mathcal{E}|{=}16$ hosts $\to |\mathcal{F}|{=}5$ features, per-agent observability mask $M^i$ as defined in Sec.~\ref{app:exp-cc4} together with the Phase~3 entity-stage mask $M^{(\text{ent})}_{j^{*}}(g^{*})$ of Sec.~\ref{sec:method-hicomm}, unicast $k{=}1$, PPO hyper-parameters, blue checkpoint) exactly as in Section~\ref{sec:exp-cc4}. The variants partition into three axis groups (\emph{payload form}, \emph{retrieval factorisation}, \emph{gradient pathway}) so that the marginal contribution of each design choice can be read off the resulting table. Each variant is trained on both IPPO and MAPPO and evaluated under the same protocol as Table~\ref{tab:summary}.

\paragraph{Module naming.} Throughout this section we use the implementation names \textsc{QueryNet}, \textsc{CommScheduler}, \textsc{DecodeNet}, and \texttt{host\_selector} for the four \textsc{HiComm} sub-modules of Sec.~\ref{sec:method-hicomm}. Concretely, \textsc{QueryNet} produces the receiver query $q_i = \tanh(f_\text{q}(h_i))$ (Phase~1); \textsc{CommScheduler} produces the Stage~2 sender-selection logits $s_{ij\mid g^{*}}=\mathrm{score}(q_i,\Phi(j,g^{*};h_i))$; \textsc{DecodeNet} jointly hosts the Stage~1 group keys $\{\kappa_g\}_{g=1}^{|\mathcal{G}|}$ (used at the receiver to pick $g^{*}$) and the Phase~3 entity projection $\psi(\cdot)$ (used at sender $j^{*}$ to pick $\ell^{*}$); and \texttt{host\_selector} is the parameter-free gather $z^{j^{*}}[g^{*},\ell^{*},:]$ that returns the addressed feature vector. Because $\kappa_g$ and $\psi$ live in \textsc{DecodeNet}, detaching ``\textsc{DecodeNet}'s subnet/host Gumbel logits'' below means detaching both the Stage~1 receiver-side group logits and the Phase~3 sender-side entity logits.

\paragraph{Payload form (observation entry vs.\ encoded).} Two variants probe the claim of Section~\ref{sec:method-hicomm} that the message should be the entry of the sender's observation tensor at the retrieved coordinate, by replacing it with reconstruction-style encoders. \textsc{HiComm-encoded} replaces the raw $|\mathcal{F}|{=}5$-dim feature-vector lookup at the queried coordinate with a learned message encoder over the sender's full host matrix (autoencoder-style, comparable in capacity to \textsc{CACOM}'s message head), so the receiver consumes a compressed summary rather than the feature vector itself; everything else is identical to \textsc{HiComm}. \textsc{HiComm-encoded-cell} keeps the two-stage hierarchical retrieval and unicast scheduler untouched but appends a small \texttt{Linear($5{\to}8$)$\to$ReLU$\to$Linear($8{\to}5$)} MLP encoder after \texttt{host\_selector}'s $5$-dim raw cell output, so the receiver consumes an encoded version of the same selected cell rather than the cell itself.

\paragraph{Retrieval factorisation (hierarchical vs.\ flat).} A single variant probes the contribution of the two-stage subnet$\to$host decode that gives \textsc{HiComm} its name. \textsc{HiComm-Flat} collapses the receiver-side Stage~1 group pick (against $\{\kappa_g\}_{g=1}^{|\mathcal{G}|}$, Sec.~\ref{sec:method-hicomm}) and the sender-side Phase~3 entity pick (against $\psi(z^{j^{*}}[g,\ell,:])$) into a single $144$-way Gumbel-Softmax over the flattened $9{\times}16=144$ (subnet,~host) coordinate set, executed at sender $j^{*}$ once $j^{*}$ has been selected by Stage~2; the receiver no longer transmits a topic group $g^{*}$ and the sender alone produces the joint coordinate $(g^{*},\ell^{*})$. Phase~1 query, Phase~2 Stage~2 sender selection, the unicast bandwidth budget, and the raw-cell payload are all unchanged, so the only quantity that changes is whether the (group,~entity) retrieval is factored hierarchically across the receiver/sender boundary or modelled jointly at the sender.

\paragraph{Gradient pathway (which modules receive PPO task-reward gradient).} The remaining variants stress-test the task-gradient pathway emphasised in Section~\ref{sec:method-hicomm} by selectively detaching each \textsc{HiComm} module's output before the actor head, during the differentiable forward. Recall that the actor objective in Sec.~\ref{sec:method-hicomm} is $\mathcal{L}_\text{actor}=\mathcal{L}^{\text{PPO}}-c_H\mathcal{H}+\lambda_{\text{ig}}(t)\,\mathcal{L}_\text{ig}$, where $\mathcal{L}_\text{ig}$ is the information-gain shaping prior cross-entropying the Stage~1, Stage~2, and Phase~3 pre-ST-GS logits against soft labels (plus a small batch-marginal entropy term that discourages sender collapse). For the ablations only, we expose its three constituent terms separately so that the contribution of each pathway is readable: $\mathcal{L}_{\text{decode}}$ is the cross-entropy on the Stage~1 subnet and Phase~3 host logits; $\mathcal{L}_{\text{target}}$ is the cross-entropy on the Stage~2 sender logits; $\mathcal{L}_{\text{explore}}$ is the batch-marginal entropy on sender selection. In the base \textsc{HiComm} all three run together as $\mathcal{L}_\text{ig}$ alongside the PPO surrogate; the experiment isolates, for each module, how much of its learning is carried by the task-reward gradient once the corresponding sub-prior becomes the sole signal.

\textsc{HiComm-NaiveComm} \texttt{.detach()}s the outputs of \textsc{CommScheduler}, \textsc{DecodeNet}, and \textsc{QueryNet} simultaneously inside \texttt{HiCommR\_Actor.forward}, so all three communication modules are trained only by the (decomposed) info-gain prior and the policy receives the resulting messages without back-propagating into the protocol. \textsc{HiComm-NoCS} \texttt{.detach()}s only \textsc{CommScheduler}'s output, leaving $\mathcal{L}_{\text{target}}+\mathcal{L}_{\text{explore}}$ as the sole training signals for target selection while the query and decode paths still receive PPO gradient. \textsc{HiComm-NoDec} \texttt{.detach()}s \textsc{DecodeNet}'s subnet/host Gumbel logits (i.e.\ both Stage~1 and Phase~3) and trains the two-stage retrieval purely from $\mathcal{L}_{\text{decode}}$. \textsc{HiComm-NoQuery} \texttt{.detach()}s \textsc{QueryNet}'s output $q$; since none of the three sub-priors is attached to \textsc{QueryNet} directly, this leaves it without any training signal and the query becomes a fixed random projection of the receiver's hidden state.

\subsubsection{Results}
\label{app:ablation-results}

Table~\ref{tab:cc4-ablation-full} reports the ablation grid on CC4 over $100$ evaluation episodes against the same fixed pre-trained blue opponent (IPPO blue policy) used in Table~\ref{tab:cc4-results-full}. Each variant is trained independently under both IPPO and MAPPO with the training budget and hyper-parameters of Section~\ref{app:exp-cc4}; only the axis listed in the variant description differs from the \textsc{HiComm} reference. We report mean episode reward, together with the number of \texttt{Impact} attempts and successes summed across the $100$ episodes. \textsc{HiComm} (top row of each block) is the reference configuration; the ordering inside each block follows the \emph{payload~$\to$~retrieval~$\to$~gradient} axis grouping of the variant descriptions above.

\begin{table}[h]
\centering
\small
\caption{CC4 ablation grid over $100$ evaluation episodes against the fixed IPPO blue opponent (red-reward convention; higher is better). Each variant differs from \textsc{HiComm} along exactly one axis; all other components, training budget, and PPO hyper-parameters are held fixed. \#Impact attempts/successes are summed over the $100$ episodes. \textbf{Bold} $=$ best reward within each algorithm block; \underline{underline} $=$ best Success~\#Impact within each algorithm block.}
\label{tab:cc4-ablation-full}
\renewcommand{\arraystretch}{1.15}
\begin{tabular}{llrrr}
\toprule
Algorithm & Variant & Reward & Total~\#Impact & Success~\#Impact \\
\midrule
IPPO  & \textsc{HiComm} (ref.)        & $\mathbf{1{,}527}$   & $49{,}499$ & $\underline{262}$  \\
IPPO  & \textsc{HiComm-encoded}       & $961$                & $41{,}088$ & $152$              \\
IPPO  & \textsc{HiComm-encoded-cell}  & $844$                & $41{,}607$ & $209$              \\
IPPO  & \textsc{HiComm-Flat}          & $1{,}381$            & $39{,}272$ & $243$              \\
IPPO  & \textsc{HiComm-NaiveComm}     & $1{,}331$            & $39{,}328$ & $198$              \\
IPPO  & \textsc{HiComm-NoCS}          & $1{,}481$            & $46{,}777$ & $245$              \\
IPPO  & \textsc{HiComm-NoDec}         & $1{,}307$            & $48{,}453$ & $216$              \\
IPPO  & \textsc{HiComm-NoQuery}       & $1{,}422$            & $48{,}032$ & $210$              \\
\midrule
MAPPO & \textsc{HiComm} (ref.)        & $\mathbf{1{,}941}$   & $78{,}216$ & $\underline{246}$  \\
MAPPO & \textsc{HiComm-encoded}       & $1{,}109$            & $40{,}993$ & $197$              \\
MAPPO & \textsc{HiComm-encoded-cell}  & $1{,}544$            & $42{,}450$ & $214$              \\
MAPPO & \textsc{HiComm-Flat}          & $690$                & $41{,}109$ & $141$              \\
MAPPO & \textsc{HiComm-NaiveComm}     & $864$                & $38{,}515$ & $198$              \\
MAPPO & \textsc{HiComm-NoCS}          & $1{,}645$            & $32{,}724$ & $223$              \\
MAPPO & \textsc{HiComm-NoDec}         & $1{,}083$            & $47{,}956$ & $170$              \\
MAPPO & \textsc{HiComm-NoQuery}       & $618$                & $40{,}268$ & $138$              \\
\bottomrule
\end{tabular}
\end{table}

Across both algorithm blocks, \textsc{HiComm} (ref.) is the unique row that is strictly best on both reward and Success~\#Impact, and every single-axis variant loses on both metrics simultaneously. The Total~\#Impact column further shows that the reward gap is not driven by attempt count: variants that lower Total~\#Impact still lose Success~\#Impact, so the regression reflects worse strategic choices rather than fewer attempts. We read the three axis groups in turn.

\paragraph{Payload form.} Both encoded variants degrade on both algorithms, and Success~\#Impact tracks the reward drop. The cell-level encoder is the milder of the two but still costly, even though it is only a small MLP appended to \texttt{host\_selector}'s raw cell output: a learned bottleneck whose reconstruction objective is decoupled from policy reward is enough to disrupt the raw lookup. The full-host autoencoder of \textsc{HiComm-encoded} is worse still, consistent with the encoder being dominated by the mostly-zero entries of an unexplored host matrix in CC4. This confirms the design choice of Section~\ref{sec:method-hicomm}: with the receiver issuing a coordinate-level query, the natural payload is the raw observation entry at that coordinate, and inserting any encoder degrades the protocol.

\paragraph{Retrieval factorisation.} \textsc{HiComm-Flat} replaces the two-stage subnet$\to$host Gumbel-Softmax decode with a single Gumbel-Softmax over the flattened entity set. Reward drops modestly under IPPO but \emph{collapses} under MAPPO, the largest single-axis regression in the table, with Success~\#Impact also dropping sharply. The hierarchical decomposition is therefore not a parametric shortcut: it injects CC4's native subnet~$\supset$~host containment as an inductive bias that turns retrieval into two short categoricals, and the centralised MAPPO critic does not recover the lost optimisation tractability on its own.

\paragraph{Gradient pathway.} Detaching all three communication modules simultaneously (\textsc{HiComm-NaiveComm}, only the decomposed info-gain prior remains) hurts both algorithms, and substantially more under MAPPO: the prior partially compensates under IPPO but cannot recover the task-gradient signal that carries credit assignment under a centralised critic. The per-module breakdown isolates which of the three modules is most reliant on PPO gradient. \textsc{HiComm-NoQuery} (no signal at all, since none of $\mathcal{L}_{\text{decode}}$, $\mathcal{L}_{\text{target}}$, $\mathcal{L}_{\text{explore}}$ attaches to \textsc{QueryNet}) suffers the most catastrophic regression on MAPPO: without any signal, the query collapses to a fixed random projection of the receiver hidden state and neither the retrieved coordinate nor the consumed payload can be aligned with what the policy actually needs. \textsc{HiComm-NoDec} (only $\mathcal{L}_{\text{decode}}$, info-gain over $(g,\ell)$) is also clearly worse than the reference, showing that an info-gain prior is a useful but insufficient substitute for task gradient through the two-stage retrieval, since informative entries are not always reward-relevant. \textsc{HiComm-NoCS} (only $\mathcal{L}_{\text{target}}+\mathcal{L}_{\text{explore}}$) is the mildest perturbation: the per-target novelty proxy is enough to keep target selection close to the reference, indicating that \emph{which} sender to query is the easiest of the three sub-problems to learn from the prior alone. Together, the gradient-pathway block establishes that the task-reward gradient is essential precisely where the prior is least informative (\textsc{QueryNet}, then \textsc{DecodeNet}), and motivates running the full \textsc{HiComm} stack inside the actor's differentiable forward as in Section~\ref{sec:method-hicomm}.

\paragraph{Takeaway.} Across the results, we found that removing the raw observation-entry payload, the hierarchical retrieval factorisation, or the task-reward gradient pathway each induces a distinct failure mode, and no alternative training signal we tested recovers the lost performance. Across all six axis perturbations and both PPO surrogates, \textsc{HiComm} (ref.) is strictly best on reward and Success~\#Impact, supporting the claim that each of the three design choices in Section~\ref{sec:method-hicomm} is necessary for the protocol as a whole.



\end{document}